\documentclass{article}

\usepackage{microtype}
\usepackage{graphicx}
\usepackage{subcaption}
\usepackage{booktabs} 

\usepackage{hyperref}

\usepackage[preprint]{icml2026}

\usepackage{amsmath}
\usepackage{amssymb}
\usepackage{mathtools}
\usepackage{amsthm}
\usepackage{multirow}
\usepackage{makecell}

\usepackage{adjustbox}

\usepackage[capitalize,noabbrev]{cleveref}

\usepackage[most]{tcolorbox}
\usepackage{xcolor}
\definecolor{boxborder}{HTML}{009400}
\definecolor{boxinner}{HTML}{F2F2F2}

\theoremstyle{plain}

\theoremstyle{definition}

\theoremstyle{remark}

\usepackage[textsize=tiny]{todonotes}

\icmltitlerunning{Not All Code Is Equal}

\newcommand{\cc}{{\textsc{{CC}}}}
\newcommand{\lloc}{{\textsc{{LLOC}}}}
\newcommand{\codenet}{{\textsc{{CodeNet}}}}
\newcommand{\instruct}{{\textsc{{Instruct}}}}

\newcommand{\python}{{\texttt{{Python}}}}
\newcommand{\typescript}{{\texttt{{TypeScript}}}}
\newcommand{\javascript}{{\texttt{{JavaScript}}}}
\newcommand{\java}{{\texttt{{Java}}}}

\newcommand{\mathfour}{{\textsc{{Math401}}}}
\newcommand{\mathfive}{{\textsc{{Math500}}}}
\newcommand{\gpqa}{{\textsc{{GPQA}}}}
\newcommand{\hle}{{\textsc{{HLE}}}}
\newcommand{\bbeh}{{\textsc{{BBEH-mini}}}}
\newcommand{\gsmk}{{\textsc{{GSM8K}}}}

\newcommand{\qwmini}{{\texttt{{Qwen2.5-3B}}}}
\newcommand{\qwsmall}{{\texttt{{Qwen2.5-7B}}}}
\newcommand{\qwmedium}{{\texttt{{Qwen2.5-14B}}}}
\newcommand{\llmini}{{\texttt{{Llama-3.2-3B}}}}
\newcommand{\llsmall}{{\texttt{{Llama-3.1-8B}}}}
\newcommand{\mistral}{{\texttt{{Mistral-7B}}}}

\newcommand{\minspl}{{\textsc{{min}}}}
\newcommand{\lowspl}{{\textsc{{low}}}}
\newcommand{\midspl}{{\textsc{{mid}}}}
\newcommand{\highspl}{{\textsc{{high}}}}
\newcommand{\maxspl}{{\textsc{{max}}}}
\newcommand{\ctrlspl}{{\textsc{{ctrl}}}}

\newcommand{\hfurl}{https://huggingface.co/datasets/itsluketwist/NotAllCodeIsEqual}
\newcommand{\hflink}{\href{\hfurl}{\hfurl}}

\begin{document}

\twocolumn[
  \icmltitle{
    Not All Code Is Equal: A Data-Centric Study of\\
    Code Complexity and LLM Reasoning
  }



  \icmlsetsymbol{equal}{*}

  \begin{icmlauthorlist}
    \icmlauthor{Lukas Twist}{kcl}
    \icmlauthor{Shu Yang}{kaust}
    \icmlauthor{Hanqi Yan}{kcl}
    \icmlauthor{Jingzhi Gong}{kcl}
    \icmlauthor{Di Wang}{kaust}
    \icmlauthor{Helen Yannakoudakis}{kcl}
    \icmlauthor{Jie M. Zhang}{kcl}
  \end{icmlauthorlist}

  \icmlaffiliation{kcl}{King's College London, United Kingdom}
  \icmlaffiliation{kaust}{King Abdullah University of Science and Technology, Saudi Arabia}

  \icmlcorrespondingauthor{Lukas Twist}{lukas.twist@kcl.ac.uk}

  \icmlkeywords{Machine Learning, ICML}

  \vskip 0.3in
]



\printAffiliationsAndNotice{}  

\begin{abstract}

Large Language Models (LLMs) increasingly exhibit strong reasoning abilities, often attributed to their capacity to generate chain-of-thought–style intermediate reasoning.
Recent work suggests that exposure to code can further enhance these skills, but existing studies largely treat code as a generic training signal, leaving open the question of which properties of code actually contribute to improved reasoning.
To address this gap, we study the structural complexity of code, which captures control flow and compositional structure that may shape how models internalise multi-step reasoning during fine-tuning.
We examine two complementary settings: \emph{solution-driven complexity}, where complexity varies across multiple solutions to the same problem, and \emph{problem-driven complexity}, where complexity reflects variation in the underlying tasks.
Using cyclomatic complexity and logical lines of code to construct controlled fine-tuning datasets, we evaluate a range of open-weight LLMs on diverse reasoning benchmarks.
Our findings show that although code can improve reasoning, structural properties strongly determine its usefulness.
In 83\% of experiments, restricting fine-tuning data to a specific structural complexity range outperforms training on structurally diverse code, pointing to a data-centric path for improving reasoning beyond scaling.

\end{abstract}


\begin{figure*}
    \centering
    \includegraphics[width=\textwidth]{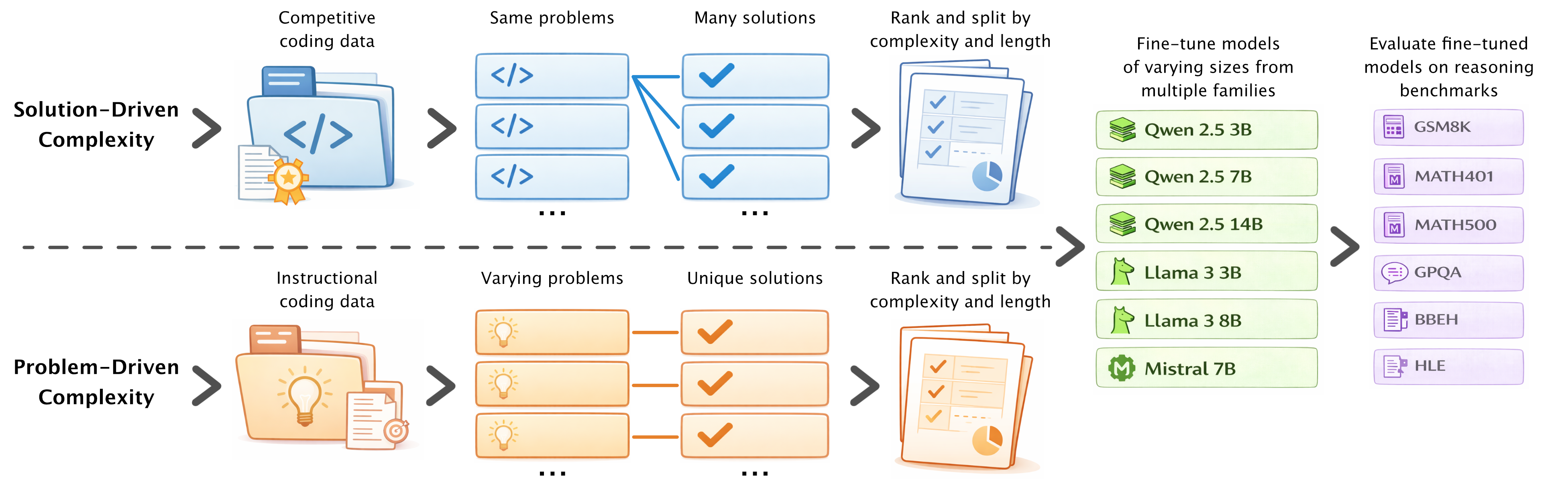}
    \caption{
        \textbf{\textit{Overview of our experimental pipeline.}}
        We construct two complementary datasets that control structural code complexity through \textit{solution-driven complexity} (top) and \textit{problem-driven complexity} (bottom), creating twelve splits for each (five different complexity levels and one control, for two different complexity metrics).
        We then use these splits to fine-tune six models from three model families and evaluate downstream reasoning performance across six widely-used reasoning benchmarks.
    }
    \label{fig:pipeline}
\end{figure*}

\section{Introduction}

Large language models (LLMs) have rapidly evolved from surface-level language processing systems into capable problem solvers, exhibiting increasingly strong reasoning behaviours across mathematical, logical, and multi-disciplinary tasks~\cite{weiEmergentAbilitiesLarge, huangReasoningLargeLanguage2023}.
A large body of work attributes these gains to the use of Chain-of-Thought (CoT) style explanations, in which models generate intermediate reasoning steps before producing an answer~\cite{weiChainofThoughtPromptingElicits2022, wangSelfConsistencyImprovesChain2023}.
CoTs have been extensively studied in natural language settings, with recent theoretical analyses providing insight into why structured intermediate traces benefit model reasoning~\cite{fengRevealingMysteryChain2023}, suggesting that symbolic scaffolding reduces search complexity and stabilises problem decomposition.

A related ``program-of-thought'' phenomenon has been observed for code, where programs provide explicit reasoning structure that would otherwise be expressed in natural language~\cite{chenProgramThoughtsPrompting2023a}.
This is because code naturally expresses control flow, branching, and intermediate computation--structures that are useful for multi-step reasoning--and has consequently been used as a structured signal to encourage CoT~\cite{linScalingCodeAssistedChainofThoughts2025}.
Beyond inference-time scaffolding, exposure to code during training has also been shown to enhance reasoning more broadly: models trained on code data often demonstrate improved multi-step reasoning and quantitative problem solving compared to models trained solely on natural language~\cite{zhangUnveilingImpactCoding2024, waheedCodeInducedReasoningLLMs2025, yangCodeThinkThink2025}.

Despite this emerging evidence, the properties of code that contribute to these gains remain under-explored.
Existing studies largely treat code as an undifferentiated training signal, without examining which characteristics of the code are actually important~\cite{aryabumiCodeNotCode2024, zhangUnveilingImpactCoding2024, waheedCodeInducedReasoningLLMs2025}.
To fill this gap, we ask whether \emph{fine-grained, measurable properties} of code can systematically shape the downstream reasoning performance of LLMs when used for fine-tuning.
We focus on \emph{structural complexity} as a candidate driver of code-induced reasoning gains, motivated by the fact that more complex programs exhibit deeper branching and richer execution paths~\cite{fentonSoftwareMetricsRigorous2014}, which can be viewed as an implicit form of a structured reasoning trace that exposes models to multi-step decomposition patterns during fine-tuning.

Specifically, we investigate this question through two complementary settings (\Cref{fig:pipeline}):
\textit{\textbf{(i)}}~\emph{Solution-driven complexity}, where complexity arises from the structure of the code itself. We use multiple solutions to identical programming problems to vary complexity while holding tasks fixed.
\textit{\textbf{(ii)}}~\emph{Problem-driven complexity}, this setting reflects variation in the difficulty of the underlying tasks, where different prompts naturally demand solutions of differing structural complexity.
Together, these settings allow us to disentangle the effects of code complexity arising from how code is written versus which problem the code is solving.
We construct two datasets with controlled complexity variation in \python, \javascript~and \java, using cyclomatic complexity~\cite{mccabeComplexityMeasure1976} and logical code lines~\cite{nguyenSLOCCountingStandard2007} as complementary structural metrics.
We then fine-tune a diverse set of open-weight models across multiple parameter scales and evaluate their reasoning performance on publicly available benchmarks spanning mathematical and multi-disciplinary problem solving.

Our results reveal that \textit{not all code is equal}.
\textit{\textbf{(1)}} Code fine-tuning does not yield uniform reasoning gains: even when using code datasets previously shown to be effective, improvements vary substantially across models and benchmarks, and depend strongly on the structural complexity of the fine-tuning data.
\textit{\textbf{(2)}} The relationship between code complexity and reasoning performance is strongly non-monotonic, with accuracy typically peaking at intermediate complexity levels and degrading for both very simple and very complex code.
\textit{\textbf{(3)}} Control datasets that mix code across all complexity levels are rarely optimal: in 83\% of experiments, restricting fine-tuning to a specific complexity range yields better reasoning performance than training on a diverse code corpus.

Our results indicate that the usefulness of code as a training signal may be governed by its structural properties.
This challenges the prevailing assumption that greater diversity or quantity of code is inherently beneficial~\cite{liuDatasetsLargeLanguage2025, abedIncreasingLLMCoding2025}, and instead points to a data-centric alternative: carefully selecting or constructing code with appropriate structural complexity matched to the specific model.
Because high-quality code data is expensive to collect and train on~\cite{chenMasteringCraftData2025}, understanding which code structures most effectively support reasoning offers a practical path to improving LLM reasoning beyond simply scaling models or datasets.

Our contributions are as follows:
\begin{itemize}
    \item We present the first systematic study of how \textit{structural properties of code}--specifically cyclomatic complexity and logical lines of code--affect the reasoning abilities of LLMs during fine-tuning.
    
    \item We provide empirical evidence that reasoning gains from code are \textit{non-uniform} and \textit{non-monotonic}, and that restricting fine-tuning data to model-specific complexity ranges often outperforms training on structurally diverse code.
    
    \item We publicly release our complexity-controlled datasets\footnote{\hflink} to support reproducibility and encourage further research on the interaction between code complexity and LLM reasoning.
\end{itemize}


\section{Related Work}

\paragraph{Reasoning abilities of LLMs.}

As LLMs evolve, reasoning and problem solving have emerged as core abilities, often appearing once models reach sufficient scale~\cite{weiEmergentAbilitiesLarge}.
Early work showed that sufficiently large models can exhibit multi-step reasoning when prompted to generate intermediate \textit{chain-of-thought}~\cite{weiChainofThoughtPromptingElicits2022}, an observation that catalysed substantial research into understanding and improving reasoning performance~\cite{fengRevealingMysteryChain2023, huangReasoningLargeLanguage2023}.
Reasoning can be improved through a range of approaches, including prompt engineering~\cite{kojimaLargeLanguageModels2022}, novel decoding strategies~\cite{wangSelfConsistencyImprovesChain2023}, fine-tuning~\cite{trungReFTReasoningReinforced2024}, and reinforcement learning~\cite{wangReinforcementLearningReasoning}.
Evaluation practices have evolved alongside these methods.
Multi-step reasoning benchmarks such as \gsmk~\cite{cobbeTrainingVerifiersSolve2021a} and \textsc{MATH}~\cite{hendrycksMeasuringMathematicalProblem2021} are widely used to assess arithmetic and logical reasoning, while more recent benchmark suites, including \textsc{BIG-Bench}~\cite{kazemiBIGBenchExtraHard2025a} and \textsc{ARC-AGI}~\cite{cholletARCAGI2NewChallenge2025}, probe a broader range of complex and abstract reasoning behaviours.

\paragraph{Using code to enhance LLM reasoning.}

Using code and code-instruction data has not only been shown to consistently improve LLM performance on code-generation and related tasks~\cite{roziereCodeLlamaOpen2024, weiMagicoderEmpoweringCode2024}, but also to improve general reasoning capabilities~\cite{yangCodeThinkThink2025}.
Recent studies investigate this phenomenon from complementary perspectives.
Some examine how the inclusion or perturbation of code data during fine-tuning affects reasoning performance~\cite{zhangUnveilingImpactCoding2024, waheedCodeInducedReasoningLLMs2025}, while others analyse the role of code earlier in the training pipeline, such as during pretraining or continued pretraining~\cite{maWhichTrainingStage2023, aryabumiCodeNotCode2024}.
A related line of work explores explicitly converting reasoning traces into code-like representations to leverage code as a structured training signal~\cite{linScalingCodeAssistedChainofThoughts2025}.
An alternative line of work explores \textit{program-of-thought} prompting, where intermediate reasoning is externalised into executable code to reduce cognitive load and improve error handling in symbolic tasks~\cite{chenProgramThoughtsPrompting2023a}.
Overall, prior work demonstrates that code can improve reasoning, but offers limited insight into which properties of code best drive these gains--an omission our work addresses by systematically varying code complexity during fine-tuning.

\paragraph{Code complexity metrics.}

The analysis of program structure via static code metrics has a long history in software engineering~\cite{fentonSoftwareMetricsRigorous2014}.
Foundational measures such as cyclomatic complexity~\cite{mccabeComplexityMeasure1976} and Halstead metrics~\cite{halsteadElementsSoftwareScience1977} quantify structural and control-flow properties of programs, while a broader ecosystem of metrics captures complementary aspects including program size~\cite{nguyenSLOCCountingStandard2007}, maintainability~\cite{303623}, and coupling or cohesion~\cite{tiwariCouplingCohesionMetrics2018}.
These classical metrics continue to play a role in modern machine learning for code.
Recent work on complexity-aware code generation uses suites of established metrics to analyse and guide LLM behaviour, showing that explicit complexity signals can improve performance~\cite{sepidbandEnhancingLLMBasedCode2025}.
Complexity metrics have also been used as discriminative features for vulnerability assessment~\cite{tehraniAssessingVulnerabilitySmart2025} and automatic defect detection~\cite{cernauUnveilingHybridCyclomatic2025}, reinforcing their utility.
Beyond these targeted applications, recent studies have evaluated LLMs’ ability to reason about code maintainability under controlled complexity conditions~\cite{dillmannEvaluationLargeLanguage2024}.
Collectively, these lines of work support viewing code complexity metrics as interpretable, quantitative signals--motivating our study of how varying complexity in fine-tuning data affects reasoning behaviour in LLMs.


\section{Methodology}

\subsection{Problem Formulation}

This work investigates whether the structural complexity of code used during fine-tuning influences the reasoning abilities of LLMs.
Prior studies have shown that exposure to code can improve reasoning~\cite{zhangUnveilingImpactCoding2024, waheedCodeInducedReasoningLLMs2025}, yet the properties of the code that drive these gains remain poorly understood.
We study this question through two complementary notions of complexity, which we define for the purposes of this work.
In \textit{solution-driven complexity}, complexity arises from the structure of the code itself: we use multiple solutions to the same underlying problems to isolate code complexity as a variable.
In contrast, \textit{problem-driven complexity} reflects the intrinsic difficulty of the task, where training data consists of a diverse set of problems that naturally require solutions of varying structural complexity.
By disentangling these two sources of complexity and controlling them through dataset construction, we aim to isolate how fine-tuning on code of varying complexity affects downstream reasoning performance.

\begin{figure}
    \centering
    \includegraphics[width=0.9\columnwidth]{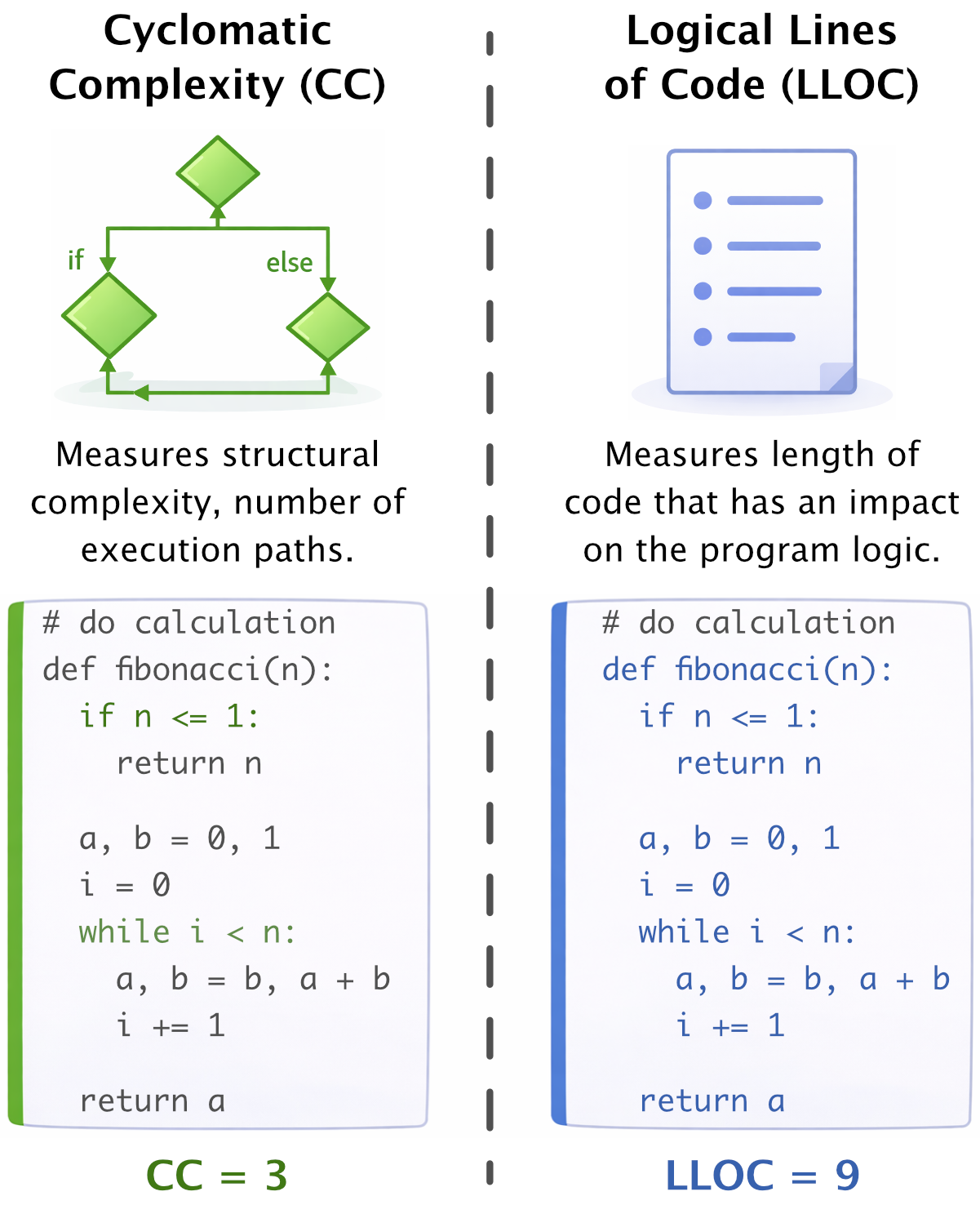}
    \caption{
        \textbf{\textit{Comparison of the code metrics used in this study.}}
        We calculate both metrics for our \textit{solution-driven complexity} dataset (\codenet) and our \textit{problem-driven complexity} dataset (\instruct).
        Together they allow us to disentangle the effects of structural complexity from size-based complexity.
    }
    \label{fig:metrics}
\end{figure}

\subsection{Code Complexity Metrics}

\paragraph{Metric selection.}
Code complexity has long been studied in software engineering through static analysis, where quantitative metrics are used to characterise structural properties of programs such as control flow and size~\cite{fentonSoftwareMetricsRigorous2014}.
In this work, we are interested in complexity measures that plausibly relate to reasoning: intuitively, code with richer control flow and more elaborate structure may expose models to patterns that resemble multi-step reasoning.
Guided by this motivation, we focus on two metrics (\Cref{fig:metrics}):
\textit{cyclomatic complexity} (\cc), which captures the number of independent execution paths through a program~\cite{mccabeComplexityMeasure1976};
and \textit{logical lines of code} (\lloc), a simpler size-based metric that reflects the amount of executable logic independent of formatting or comments~\cite{nguyenSLOCCountingStandard2007}.
Because \cc\ tends to increase with code length, incorporating \lloc\ enables us to control for length and isolate the effect of structural complexity.

\paragraph{Metric calculation.}
We construct multilingual training datasets to capture a diverse programming problems and solution patterns, focussing on the three most used programming languages on GitHub: \python, \javascript\ (including \typescript), and \java~\cite{githubstaffOctoverseNewDeveloper2025}.
To ensure consistency and reproducibility across languages, we rely on established open-source static analysis tools for metric computation.
For \python, we use \texttt{Radon}~\cite{rubikRubikRadon};
for \javascript, we use \texttt{escomplex}~\cite{escomplexEscomplexEscomplex};
and for \java, we use \texttt{PMD}~\cite{pmdPmdPmd}.
All tools are applied using a uniform preprocessing pipeline.
\textit{Implementation details are provided in Appendix~\ref{app:detail_complexity}.}

\subsection{Dataset Construction}

\paragraph{Solution-driven complexity.}
We require a dataset containing many distinct code solutions to the same underlying problems, allowing complexity to vary independently of problem semantics.
We therefore adopt Project CodeNet~\cite{puriCodeNetLargeScaleAI2021a}, a large-scale corpus of competitive programming problems paired with thousands of accepted solutions across multiple programming languages.
CodeNet is well suited to our setting, as it provides diverse, independently authored solutions to identical problem statements, enabling controlled comparisons of code complexity while holding the task fixed.
CodeNet problem descriptions are provided as HTML, and solutions are presented as standalone code snippets, neither of which is directly suitable for instruction-style fine-tuning.
We therefore augment the dataset, using an LLM to convert HTML problem statements into concise natural language instructions and to wrap each solution in a consistent response format.
We use the \texttt{gpt-5-mini-2025-08-07} model for this augmentation step, as it is a cost-effective state-of-the-art model that performs reliably on well-defined tasks~\cite{openaiGPT5MiniAPI2025}.
\textit{Augmentation prompts are detailed in Appendix~\ref{app:augment}.}

For each problem, we compute \cc\ and \lloc\ over all available \python, \javascript, and \java\ solutions.
For both metrics, we split the dataset by ranking solutions within each problem--language pair and selecting five representative solutions: the least complex, the most complex, and three evenly spaced solutions from the remainder of the distribution.
These selections form five complexity-controlled dataset splits corresponding to increasing levels of solution-driven complexity (splits named: \minspl, \lowspl, \midspl, \highspl, \maxspl).
In addition, we construct a control dataset (\ctrlspl) by sampling solutions uniformly across these complexity levels.
The resulting \codenet\ dataset comprises twelve splits (six per metric) of 8,087 samples.

\paragraph{Problem-driven complexity.}
Next, we construct a dataset in which variation in code complexity primarily reflects differences in the underlying tasks rather than alternative solutions to the same problem.
To do this, we gather three high-quality instruction--response code datasets--Magicoder~\cite{ise-uiucIseuiucMagicoderEvolInstruct110K2023}, Evol-Instruct~\cite{nickroshNickroshEvolInstructCode80kV12024}, and WizardLM~\cite{rombodawgRombodawgCode_instruct_alpaca_vicuna_wizardlm_56k_backup}--which have been previously used for training LLMs~\cite{weiMagicoderEmpoweringCode2024} and improving their reasoning~\cite{waheedCodeInducedReasoningLLMs2025}.
Unlike CodeNet, these datasets pair natural language instructions with a single reference solution, and thus reflect settings where task difficulty and solution complexity are intrinsically linked.

For each dataset response, we use regular expressions to extract code blocks and identify the programming language.
We retain only samples containing \python, \javascript, or \java\ code and compute \cc\ and \lloc\ for each.
After filtering, the dataset contains 77,686 \python, 13,949 \javascript\ and 8,054 \java\ samples.
For each metric, we rank samples independently within each language and partition them into five disjoint complexity bins.
To construct balanced datasets comparable in size to \codenet, we include all available \javascript\ and \java\ samples and supplement them with \python\ samples until each split contains 8,087 samples (splits named: \textsc{min}, \lowspl, \midspl, \highspl, \maxspl).
In addition, we construct a control dataset (\ctrlspl) that samples across languages and complexity levels.
The resulting \instruct\ dataset again comprises twelve splits (six per metric), enabling controlled comparisons with \codenet.

\paragraph{Natural language baseline.}
To disentangle the effects of code exposure from general fine-tuning, we include a natural language (NL) baseline, following the methodology of prior work~\cite{zhangUnveilingImpactCoding2024}.
Specifically, we reuse the same non-code ShareGPT dataset employed in that study, as it is comparable in scale to our code datasets and enables a controlled comparison.
From this corpus, we sample 8,087 records to match the exact size of each code-based split.
The resulting dataset therefore isolates the effect of fine-tuning itself, which is important because fine-tuning alone has been shown to sometimes induce non-trivial shifts in downstream reasoning behaviour~\cite{luoEmpiricalStudyCatastrophic2025}.
This allows us to attribute any observed reasoning changes specifically to properties of the code rather than to fine-tuning alone.

\paragraph{}
\textit{Full dataset statistics are available in Appendix~\ref{app:datasets}, and the final datasets are publicly available on Hugging Face\footnote{\hflink}.}

\subsection{Evaluation Set-up}

\paragraph{Model selection.}
We evaluate our approach across a diverse set of models to assess the impact of code complexity more broadly.
Specifically, we select models spanning multiple sizes from three major families: \texttt{Qwen-2.5} \texttt{3B}, \texttt{7B} and \texttt{14B}~\cite{qwenQwen25TechnicalReport2025}; \texttt{Llama-3} \texttt{3B} and \texttt{8B}~\cite{grattafioriLlama3Herd2024}; and \mistral~\cite{jiangMistral7B2023}.
These families were chosen due to their open availability, their widespread use in academic evaluation, and their coverage of similar parameter scales (\texttt{3B}--\texttt{14B}), allowing us to study whether complexity-related effects generalise across model sizes and architectures.

\paragraph{Training configurations.}
All models are fine-tuned for two epochs over each dataset split using LoRA~\cite{huLoRALowRankAdaptation2021a} with a learning rate of $2\times10^{-5}$, AdamW optimisation, and a cosine learning rate schedule with a warm-up ratio of $0.1$.
Training is performed on NVIDIA A100 GPUs, and all experiments use the same optimiser, scheduling, and adaptation configuration to ensure comparability.

\paragraph{Evaluation benchmarks.}
To measure downstream reasoning performance, we evaluate on six publicly available benchmarks that cover a broad spectrum of reasoning demands.
We first consider math-focused benchmarks that emphasise multi-step computation and logical reasoning:
\gsmk, a widely used dataset of grade-school arithmetic problems~\cite{cobbeTrainingVerifiersSolve2021a};
\mathfour, which contains 401 arithmetic reasoning problems~\cite{yuanHowWellLarge2023};
and \mathfive, comprising 500 mathematical problems covering a broader range of topics and difficulty levels~\cite{lightmanLetsVerifyStep2023}.
Beyond mathematics, \gpqa\ evaluates graduate-level quantitative reasoning, including physics-based problem solving~\cite{reinGPQAGraduateLevelGoogleProof2023};
\bbeh\ is a curated subset of \textsc{BIG-Bench Extra Hard} designed to probe complex and diverse reasoning behaviours~\cite{kazemiBIGBenchExtraHard2025a};
and \hle\ (Humanity’s Last Exam) spans multi-disciplinary questions across the humanities, sciences, and general knowledge domains~\cite{phanHumanitysLastExam2025}.

\paragraph{}
\textit{Further evaluation details are available in Appendix~\ref{app:models}.}



\begin{figure*}[ht]
    \centering
    \includegraphics[width=\textwidth]{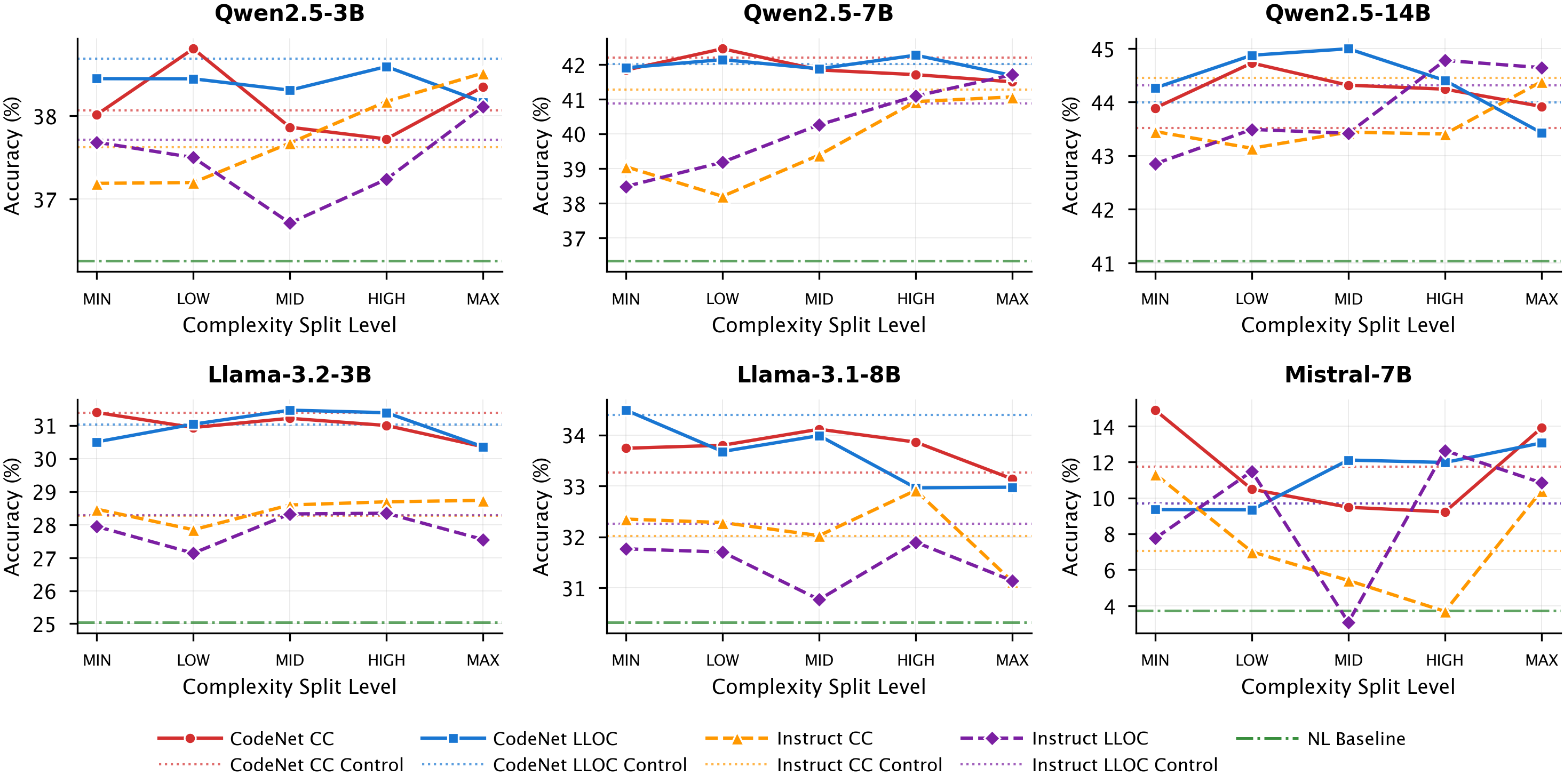}
    
    \caption{
        \textbf{\textit{Per-model reasoning performance across complexity splits.}}
        Reasoning accuracy for each model after fine-tuning on complexity-controlled dataset splits.
        Solid lines correspond to solution-driven complexity splits (\codenet), dashed lines to problem-driven complexity splits (\instruct);
        dotted horizontal lines indicate the corresponding results for control (\ctrlspl) datasets with mixed complexity;
        the dash-dotted green line denotes the model's natural language (NL) baseline after fine-tuning on a strictly non-code dataset.
    }

    \label{fig:mainresult}
\end{figure*}

\begin{figure*}[ht]
    \centering
    \includegraphics[width=\textwidth]{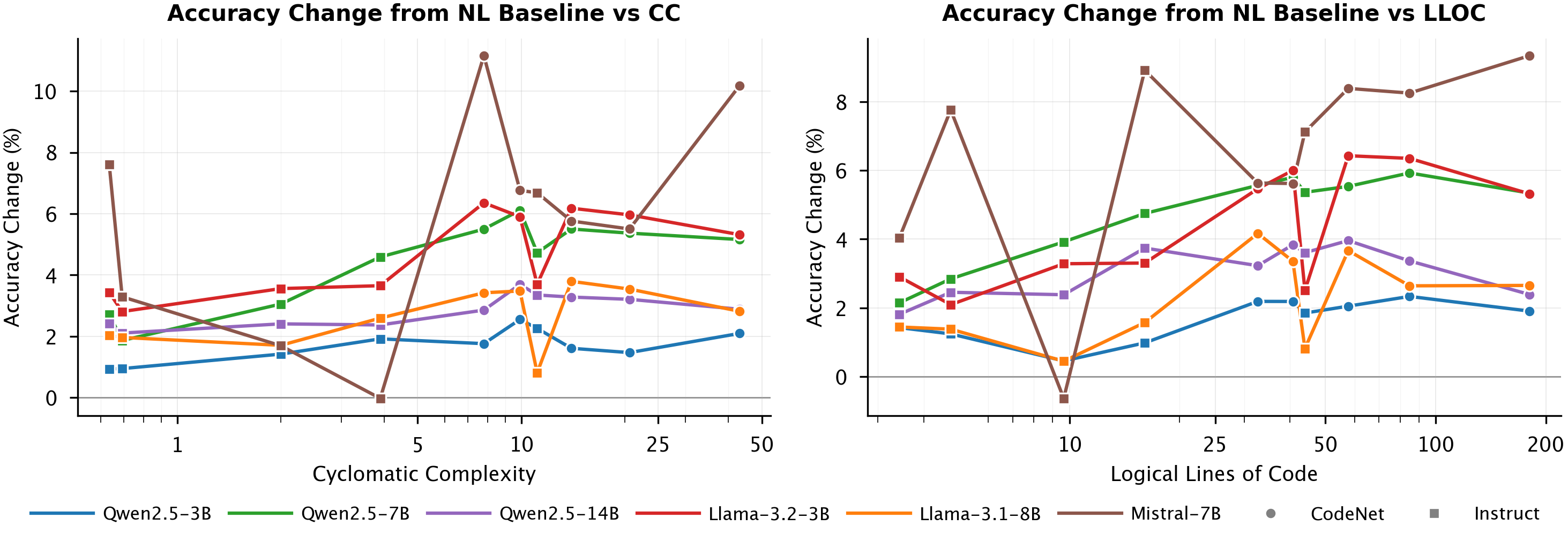}

    \caption{
        \textbf{\textit{Average reasoning change vs. code complexity.}}
        Average accuracy change compared to the NL baseline across all six reasoning benchmarks as a function of cyclomatic complexity (\cc, left) and logical lines of code (\lloc, right).
        Results are shown as a single line that includes both the solution-driven (\codenet; circles) and the problem-driven (\instruct; squares) complexity datasets.
    }

    \label{fig:accuracyvsmetric}
\end{figure*}

\section{Results}

We begin by presenting high-level findings that are robust across models, datasets, and benchmarks.
We then turn to a more fine-grained analysis that unpacks \emph{why} these patterns arise, and how they differ across model families, dataset constructions, and reasoning tasks.
\textit{Appendix~\ref{app:results} reports complete results for all models and benchmark datasets.}

\subsection{Main Findings}
\label{sec:main_results}

\paragraph{Code fine-tuning yields non-uniform reasoning gains.}
\Cref{fig:mainresult} shows average reasoning accuracy across six benchmarks after fine-tuning on code data split by cyclomatic complexity (\cc) or logical lines of code (\lloc), under both the solution-driven (\codenet) and problem-driven (\instruct) settings.
Across models, fine-tuning on code often improves reasoning relative to the natural language (NL) baseline; however, these gains are clearly \emph{not uniform}.
The same model can benefit substantially from one code subset, yet show negligible--or even negative--changes when fine-tuned on code of a different structural complexity.

All experiments use small, tightly controlled fine-tuning datasets (8,087 samples per split) to ensure strict comparability across complexity levels.
Under this regime, code--even when drawn from datasets previously shown to support reasoning~\cite{yangCodeThinkThink2025}--is not a guaranteed source of improvement across models or benchmarks.
Instead, performance exhibits pronounced sensitivity to structural complexity, with sharp peaks and troughs across adjacent splits.
Taken together, these results show that the effectiveness of code fine-tuning depends critically on \emph{which} code is used, rather than merely on the presence of code itself.

\paragraph{Reasoning gains are non-monotonic and peak at intermediate complexity.}
Across all model families, the relationship between code complexity and downstream reasoning performance is strongly \emph{non-monotonic}.
Rather than improving steadily as complexity increases, accuracy curves typically exhibit a peak at intermediate levels of both \cc and \lloc\ (\Cref{fig:mainresult,fig:accuracyvsmetric}), a trend that is visible across models of all sizes.
Importantly, this pattern reflects a relative trade-off rather than a uniformly positive effect of increasing complexity.
Both very simple and very complex code tend to underperform compared to mid-range splits, indicating that structural complexity must fall within a narrow, model-dependent range to be most beneficial.

\paragraph{Mixed-complexity control datasets are rarely optimal.}
A striking and consistent finding is that control datasets--constructed by uniformly mixing code across all complexity levels--are almost never optimal.
Across 20 of the 24 model–dataset combinations, at least one restricted complexity split outperforms its corresponding control (\Cref{fig:mainresult}).

This finding directly challenges the common assumption that diversity in code corpora is inherently beneficial.
Instead, our results suggest that \emph{targeted restriction to a specific complexity range} often yields better reasoning performance than broad mixing.
Notably, this effect holds across both solution-driven and problem-driven settings, indicating that it is not an artefact of dataset construction but a general property of how models respond to structured code.

\begin{tcolorbox}[
  colback=boxinner,
  colframe=boxborder,
  coltitle=white,
  title=\textbf{Summary of main findings.},
  fonttitle=\bfseries,
  arc=4pt,
  boxrule=0.8pt,
  left=6pt,
  right=6pt,
  top=6pt,
  bottom=6pt
]
Overall, these results identify structural code complexity as a critical--and previously under-explored--factor in code-based fine-tuning for reasoning.
The central takeaway is simple but consequential: fine-tuning on code does not guarantee reasoning gains, and restricting training data to an appropriate complexity or length range can yield stronger improvements than training on a large, mixed code corpus.
\end{tcolorbox}


\subsection{Fine-Grained Analysis Across Models and Datasets}

\begin{figure*}[ht]
    \centering
    \includegraphics[width=\textwidth]{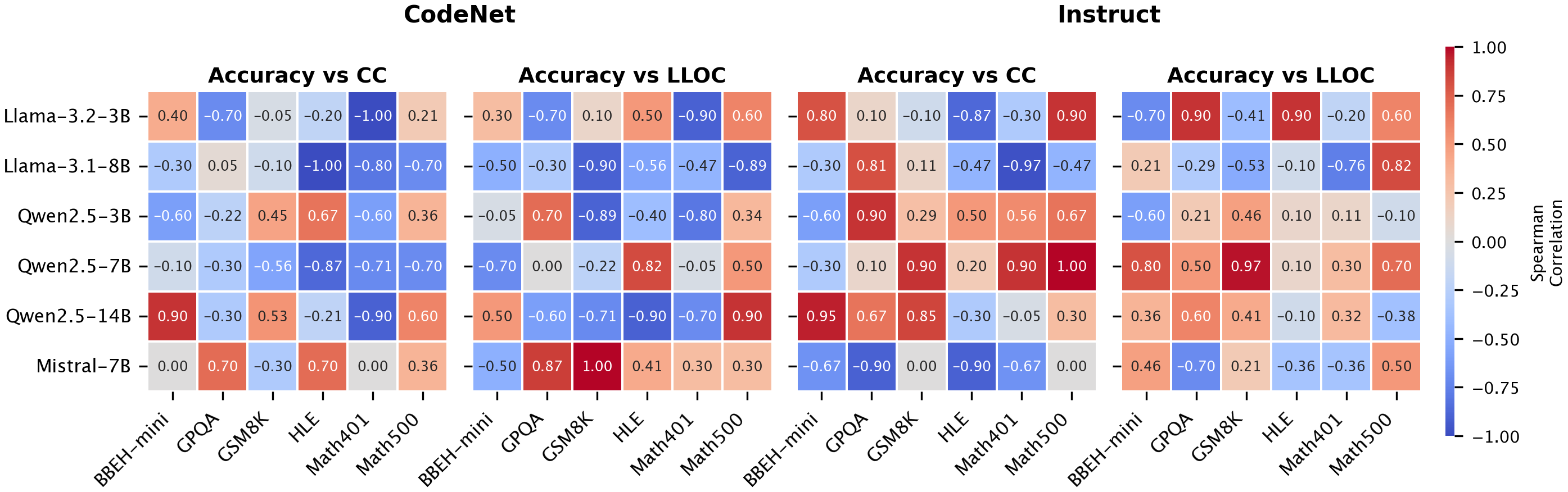}
    
    \caption{
        \textbf{\textit{Correlation between code complexity and downstream reasoning accuracy.}}
        Spearman correlations between benchmark accuracy and training-data complexity level, computed across complexity-controlled dataset splits.
        Results are shown separately for solution-driven (\codenet) and problem-driven (\instruct) settings, using cyclomatic complexity (CC) and logical lines of code (LLOC) as structural measures.
        \textit{Correlation calculation is detailed in Appendix~\ref{app:correlation}.}
    }

    \label{fig:correlation}
\end{figure*}

\paragraph{Problem-driven complexity shows stronger and more consistent effects than solution-driven complexity.}
\Cref{fig:correlation} reports Spearman correlations between training-data complexity and downstream benchmark accuracy.
Across both \cc\ and \lloc, \instruct\ exhibits more consistent positive correlations than \codenet.
This suggests that exposure to problems that \emph{require} more complex solutions is more reliably beneficial for reasoning than simply training on arbitrarily complex code.
This pattern is particularly pronounced for the \texttt{Qwen} family, which shows predominantly positive correlations across multiple benchmarks under the \instruct\ setting.
Consistent with this observation, \Cref{fig:mainresult} shows clearer upward trends in the \instruct\ accuracy curves, especially for \qwsmall.

\paragraph{Cyclomatic complexity provides a more reliable and interpretable signal than logical lines of code.}
Although both \cc\ and \lloc\ capture aspects of structural complexity, their effects on reasoning differ in stability and interpretability.
Across models and datasets, performance trends with respect to \cc\ are generally smoother and more consistent than those observed for \lloc.
In contrast, \lloc-based splits often exhibit sharper fluctuations and less regular behaviour, particularly at higher complexity levels.
This difference likely reflects the nature of the metrics themselves.
\cc\ directly captures branching structure and control flow, which are closely related to multi-step reasoning processes.
\lloc, by comparison, is a coarser proxy that conflates structure with verbosity.

\paragraph{Absolute complexity matters more than dataset construction.}
For \codenet, all \texttt{Qwen} models tend to peak at the \lowspl\ \cc\ split, whereas for \instruct, they peak at the \maxspl\ \cc\ split.
Despite this apparent discrepancy, both peaks correspond to a similar absolute \cc\ value ($\approx10$; \Cref{fig:accuracyvsmetric}).
This alignment suggests that absolute structural complexity--rather than whether that complexity arises from problem difficulty or solution variation--is the dominant factor governing the impact on reasoning.
Dataset construction primarily determines \textit{how} models encounter this complexity, but the effective complexity range itself appears to be model-specific and largely invariant across settings.

\paragraph{High structural complexity can actively harm reasoning.}
Beyond being merely suboptimal, high structural complexity can in some cases actively degrade reasoning performance.
In several settings, fine-tuning on higher-complexity code can reliably reduce accuracy relative not only to other complexity ranges, but also to the natural-language (NL) baseline.
For example, the \texttt{Llama} models see multiple near-perfect negative correlations ($\rho\approx-1.00$) for \cc\ splits in both \codenet\ and \instruct\ settings (\Cref{fig:correlation}), indicating that increasing problem complexity can directly impair logical reasoning.
These effects are also visible in the accuracy curves, which commonly dip at the highest complexity levels across models (\Cref{fig:accuracyvsmetric}).
These cases highlight that structural complexity is not merely a diminishing-returns signal, but can induce negative transfer when pushed beyond a model’s effective range.

\paragraph{Mistral exhibits qualitatively different behaviour.}
Across both \cc-based datasets, \mistral\ displays a distinctive ``U''-shaped performance curve (\Cref{fig:mainresult}), benefiting most from either very simple or very complex code, with degraded performance at intermediate levels.
This behaviour suggests that \mistral\ may interact with structural complexity in a fundamentally different way to other models, potentially reflecting differences in its architecture or training processes.
We leave a deeper investigation of these effects to future work.


\section{Discussion}

Our results suggest that the relationship between code and reasoning is more structured--and more fragile--than is often assumed.
Rather than acting as a uniformly beneficial training signal, low levels of code data appear to support reasoning only when the code’s structural properties align with what a model can effectively internalise.
Below, we briefly discuss what this implies for how code helps reasoning, and when it may fail to do so.

\paragraph{Why more code is not necessarily better.}
Previous work largely assumes that increasing the amount or diversity of code data will improve downstream reasoning performance~\cite{aryabumiCodeNotCode2024, roziereCodeLlamaOpen2024, weiMagicoderEmpoweringCode2024}.
Our findings challenge this view: once structural complexity is controlled, additional or more complex code does not reliably lead to better reasoning, and can in some cases be actively harmful.
This suggests that previously reported gains from large code corpora may stem less from diversity itself, and more from incidental exposure to particular structural properties that happen to suit a given model.

\paragraph{Structural complexity as implicit code chain-of-thought.}
Both natural-language chain-of-thought prompting~\cite{weiChainofThoughtPromptingElicits2022, fengRevealingMysteryChain2023} and similar program-based techniques~\cite{chenProgramThoughtsPrompting2023a, linScalingCodeAssistedChainofThoughts2025} improve reasoning by externalising intermediate structure and decomposing problems into explicit steps.
This provides a natural way to interpret our results, where complex code supplies a similar scaffold through control flow and branching during fine-tuning.
However, when structural complexity becomes too high, this scaffold can break down, introducing brittle control flow or optimisation difficulty that obscures rather than clarifies the reasoning signal.

\paragraph{Implications for data-centric reasoning improvement.}
Rather than relying on broad, mixed corpora, carefully selecting or constructing code with appropriate structural properties can yield stronger reasoning gains at fixed data or training budgets.
This complements recent arguments that data quality and structure can rival or exceed gains from scale alone~\cite{longprePretrainersGuideTraining2024}, and is especially relevant in settings where high-quality code data is expensive to curate or fine-tune on~\cite{chenMasteringCraftData2025}.
More broadly, our results suggest that improving reasoning through code is less about increasing exposure to programming in general, and more about identifying which computational structures best support multi-step reasoning in practice.


\section{Limitations}

This study has several limitations.
First, our experiments focus on a small set of widely used programming languages and open-weight models; although these cover diverse coding styles, parameter scales, and model architectures, our findings may not fully generalise to other languages or proprietary models.
Second, we characterise structure using two established static metrics--cyclomatic complexity and logical lines of code--which capture important aspects of control flow and program size; whilst well suited to an initial investigation, they do not exhaust the space of code properties that may influence reasoning.
Finally, we study fine-tuning on relatively small datasets (8,087 samples per split) to enable controlled comparisons; while this setting reveals clear trends, effects may differ at larger scales.


\section{Conclusion}

In this work, we studied how the structural properties of the code used during fine-tuning influence the reasoning abilities of LLMs.
By systematically varying code complexity and length in both solution-driven and problem-driven settings, and evaluating across multiple model families and reasoning benchmarks, we show that code is not a uniform training signal for reasoning.
Additionally, in 83\% of experiments, restricting the fine-tuning data to an appropriate complexity or length range yields better downstream reasoning performance than training on a diverse mix of code.
Our results highlight a data-centric path for improving reasoning--one that focusses on the \textit{specific structural properties} of the code that is used for training.

Our study deliberately focusses on simple, interpretable structural metrics to enable controlled analysis, but this choice also points to several promising directions for future work.
More expressive measures of code structure--such as hybrid metrics that combine control flow, data flow, and semantic patterns, or qualitative indicators of programming techniques of varying difficulties--may better capture the aspects of code that support reasoning.
Exploring these directions may further clarify how code functions as an implicit form of chain-of-thought during training, and help translate structural insights into more targeted and efficient training data curation.


\newpage

\section*{Impact Statement}

This paper presents work whose goal is to advance the field of Machine
Learning. There are many potential societal consequences of our work, none
which we feel must be specifically highlighted here.

\bibliography{main}

@misc{waheedCodeInducedReasoningLLMs2025,
  title = {On {{Code-Induced Reasoning}} in {{LLMs}}},
  author = {Waheed, Abdul and Wu, Zhen and Ros{\'e}, Carolyn and Ippolito, Daphne},
  year = 2025,
  month = oct,
  number = {arXiv:2509.21499},
  eprint = {2509.21499},
  primaryclass = {cs},
  publisher = {arXiv},
  doi = {10.48550/arXiv.2509.21499},
  urldate = {2025-11-28},
  archiveprefix = {arXiv}
}

@inproceedings{zhangUnveilingImpactCoding2024,
  title = {Unveiling the {{Impact}} of {{Coding Data Instruction Fine-Tuning}} on {{Large Language Models Reasoning}}},
  booktitle = {Proceedings of the {{Thirty-Ninth AAAI Conference}} on {{Artificial Intelligence}} and {{Thirty-Seventh Conference}} on {{Innovative Applications}} of {{Artificial Intelligence}} and {{Fifteenth Symposium}} on {{Educational Advances}} in {{Artificial Intelligence}}},
  author = {Zhang, Xinlu and Chen, Zhiyu Zoey and Ye, Xi and Yang, Xianjun and Chen, Lichang and Wang, William Yang and Petzold, Linda Ruth},
  year = 2024,
  month = dec,
  eprint = {2405.20535},
  primaryclass = {cs},
  publisher = {arXiv},
  doi = {10.1609/aaai.v39i24.34789},
  urldate = {2025-10-16},
  archiveprefix = {arXiv}
}

@book{fentonSoftwareMetricsRigorous2014,
  title = {Software {{Metrics}}: {{A Rigorous}} and {{Practical Approach}}, {{Third Edition}}},
  shorttitle = {Software {{Metrics}}},
  author = {Fenton, Norman and Bieman, James},
  year = 2014,
  month = sep,
  edition = {3rd},
  publisher = {CRC Press, Inc.},
  address = {USA},
  isbn = {978-1-4398-3822-8}
}

@inproceedings{mccabeComplexityMeasure1976,
  title = {A Complexity Measure},
  booktitle = {Proceedings of the 2nd International Conference on {{Software}} Engineering},
  author = {McCabe, Thomas J.},
  year = 1976,
  month = oct,
  series = {{{ICSE}} '76},
  pages = {407},
  publisher = {IEEE Computer Society Press},
  address = {Washington, DC, USA},
  urldate = {2026-01-06}
}

@misc{nguyenSLOCCountingStandard2007,
  title = {A {{SLOC Counting Standard}}},
  author = {Nguyen, Vu and {Deeds-Rubin}, Sophia and Tan, Thomas and Boehm, Barry W.},
  year = 2007,
  urldate = {2026-01-06}
}

@misc{escomplexEscomplexEscomplex,
  title = {Escomplex/Escomplex},
  author = {{escomplex}},
  urldate = {2026-01-06},
  year = 2025,
  howpublished = {https://github.com/escomplex/escomplex}
}

@misc{githubstaffOctoverseNewDeveloper2025,
  title = {Octoverse: {{A}} New Developer Joins {{GitHub}} Every Second as {{AI}} Leads {{TypeScript}} to \#1},
  shorttitle = {Octoverse},
  author = {{GitHub Staff}},
  year = 2025,
  month = oct,
  journal = {The GitHub Blog},
  urldate = {2026-01-06},
  langid = {american}
}

@misc{pmdPmdPmd,
  title = {Pmd/Pmd},
  author = {{pmd}},
  journal = {GitHub},
  urldate = {2026-01-06},
  year = 2025,
  howpublished = {https://github.com/pmd/pmd/releases},
  langid = {english}
}

@misc{rubikRubikRadon,
  title = {Rubik/Radon},
  author = {{rubik}},
  urldate = {2026-01-06},
  year = 2025,
  howpublished = {https://github.com/rubik/radon}
}

@misc{puriCodeNetLargeScaleAI2021a,
  title = {{{CodeNet}}: {{A Large-Scale AI}} for {{Code Dataset}} for {{Learning}} a {{Diversity}} of {{Coding Tasks}}},
  shorttitle = {{{CodeNet}}},
  author = {Puri, Ruchir and Kung, David S. and Janssen, Geert and Zhang, Wei and Domeniconi, Giacomo and Zolotov, Vladimir and Dolby, Julian and Chen, Jie and Choudhury, Mihir and Decker, Lindsey and Thost, Veronika and Buratti, Luca and Pujar, Saurabh and Ramji, Shyam and Finkler, Ulrich and Malaika, Susan and Reiss, Frederick},
  year = 2021,
  month = aug,
  number = {arXiv:2105.12655},
  eprint = {2105.12655},
  primaryclass = {cs},
  publisher = {arXiv},
  doi = {10.48550/arXiv.2105.12655},
  urldate = {2026-01-06},
  archiveprefix = {arXiv}
}

@misc{openaiGPT5MiniAPI2025,
  title = {{{GPT-5}} Mini - {{API}}},
  author = {OpenAI},
  year = 2025,
  journal = {OpenAI API},
  urldate = {2025-08-22},
  howpublished = {https://platform.openai.com},
  langid = {american}
}

@misc{ise-uiucIseuiucMagicoderEvolInstruct110K2023,
  title = {Ise-Uiuc/{{Magicoder-Evol-Instruct-110K}}},
  author = {{ise-uiuc}},
  year = 2023,
  month = dec,
  journal = {Hugging Face},
  urldate = {2026-01-06},
  howpublished = {https://huggingface.co/datasets/ise-uiuc/Magicoder-Evol-Instruct-110K}
}

@misc{nickroshNickroshEvolInstructCode80kV12024,
  title = {Nickrosh/{{Evol-Instruct-Code-80k-v1}}},
  author = {{nickrosh}},
  year = 2024,
  month = oct,
  journal = {Hugging Face},
  urldate = {2026-01-06},
  howpublished = {https://huggingface.co/datasets/nickrosh/Evol-Instruct-Code-80k-v1}
}

@misc{rombodawgRombodawgCode_instruct_alpaca_vicuna_wizardlm_56k_backup,
  author       = {{rombodawg}},
  title        = {code\_instruct\_alpaca\_vicuna\_wizardlm\_56k\_backup},
  year         = {2023},
  howpublished = {\url{https://huggingface.co/datasets/rombodawg/code_instruct_alpaca_vicuna_wizardlm_56k_backup}}
}

@inproceedings{weiMagicoderEmpoweringCode2024,
  title = {Magicoder: Empowering Code Generation with {{OSS-INSTRUCT}}},
  shorttitle = {Magicoder},
  booktitle = {Proceedings of the 41st {{International Conference}} on {{Machine Learning}}},
  author = {Wei, Yuxiang and Wang, Zhe and Liu, Jiawei and Ding, Yifeng and Zhang, Lingming},
  year = 2024,
  month = jul,
  series = {{{ICML}}'24},
  volume = {235},
  pages = {52632--52657},
  publisher = {JMLR.org},
  address = {Vienna, Austria},
  urldate = {2026-01-06}
}

@book{halsteadElementsSoftwareScience1977,
  title = {Elements of {{Software Science}} ({{Operating}} and Programming Systems Series)},
  author = {Halstead, Maurice H.},
  year = 1977,
  month = apr,
  publisher = {Elsevier Science Inc.},
  address = {USA},
  isbn = {978-0-444-00205-1}
}

@article{303623,
  title = {Using Metrics to Evaluate Software System Maintainability},
  author = {Coleman, D. and Ash, D. and Lowther, B. and Oman, P.},
  year = 1994,
  journal = {Computer},
  volume = {27},
  number = {8},
  pages = {44--49},
  doi = {10.1109/2.303623}
}

@inproceedings{tiwariCouplingCohesionMetrics2018,
  title = {Coupling and {{Cohesion Metrics}} for {{Object-Oriented Software}}: {{A Systematic Mapping Study}}},
  shorttitle = {Coupling and {{Cohesion Metrics}} for {{Object-Oriented Software}}},
  booktitle = {Proceedings of the 11th {{Innovations}} in {{Software Engineering Conference}}},
  author = {Tiwari, Saurabh and Rathore, Santosh Singh},
  year = 2018,
  month = feb,
  series = {{{ISEC}} '18},
  pages = {1--11},
  publisher = {Association for Computing Machinery},
  address = {New York, NY, USA},
  doi = {10.1145/3172871.3172878},
  urldate = {2026-01-07},
  isbn = {978-1-4503-6398-3}
}

@misc{cernauUnveilingHybridCyclomatic2025,
  title = {Unveiling {{Hybrid Cyclomatic Complexity}}: {{A Comprehensive Analysis}} and {{Evaluation}} as an {{Integral Feature}} in {{Automatic Defect Prediction Models}}},
  shorttitle = {Unveiling {{Hybrid Cyclomatic Complexity}}},
  author = {Cernau, Laura Diana and Diosan, Laura and Serban, Camelia},
  year = 2025,
  month = apr,
  number = {arXiv:2504.00477},
  eprint = {2504.00477},
  primaryclass = {cs},
  publisher = {arXiv},
  doi = {10.48550/arXiv.2504.00477},
  urldate = {2026-01-07},
  archiveprefix = {arXiv}
}

@inproceedings{sepidbandEnhancingLLMBasedCode2025,
  title = {Enhancing {{LLM-Based Code Generation}} with {{Complexity Metrics}}: {{A Feedback-Driven Approach}}},
  shorttitle = {Enhancing {{LLM-Based Code Generation}} with {{Complexity Metrics}}},
  booktitle = {2025 {{IEEE}} 49th {{Annual Computers}}, {{Software}}, and {{Applications Conference}} ({{COMPSAC}})},
  author = {Sepidband, Melika and Taherkhani, Hamed and Wang, Song and Hemmati, Hadi},
  year = 2025,
  month = jul,
  pages = {1416--1426},
  publisher = {IEEE Computer Society},
  doi = {10.1109/COMPSAC65507.2025.00178},
  urldate = {2026-01-07},
  isbn = {979-8-3315-7434-5},
  langid = {english}
}

@misc{tehraniAssessingVulnerabilitySmart2025,
  title = {Assessing {{Vulnerability}} in {{Smart Contracts}}: {{The Role}} of {{Code Complexity Metrics}} in {{Security Analysis}}},
  shorttitle = {Assessing {{Vulnerability}} in {{Smart Contracts}}},
  author = {Tehrani, Masoud Jamshidiyan and Hashemi, Sattar},
  year = 2025,
  month = mar,
  number = {arXiv:2411.17343},
  eprint = {2411.17343},
  primaryclass = {cs},
  publisher = {arXiv},
  doi = {10.48550/arXiv.2411.17343},
  urldate = {2026-01-07},
  archiveprefix = {arXiv}
}

@misc{dillmannEvaluationLargeLanguage2024,
  title = {Evaluation of Large Language Models for Assessing Code Maintainability},
  author = {Dillmann, Marc and Siebert, Julien and Trendowicz, Adam},
  year = 2024,
  month = jan,
  number = {arXiv:2401.12714},
  eprint = {2401.12714},
  primaryclass = {cs},
  publisher = {arXiv},
  doi = {10.48550/arXiv.2401.12714},
  urldate = {2026-01-07},
  archiveprefix = {arXiv}
}

@inproceedings{huangReasoningLargeLanguage2023,
  title = {Towards {{Reasoning}} in {{Large Language Models}}: {{A Survey}}},
  shorttitle = {Towards {{Reasoning}} in {{Large Language Models}}},
  booktitle = {Findings of the {{Association}} for {{Computational Linguistics}}: {{ACL}} 2023},
  author = {Huang, Jie and Chang, Kevin Chen-Chuan},
  editor = {Rogers, Anna and {Boyd-Graber}, Jordan and Okazaki, Naoaki},
  year = 2023,
  month = jul,
  pages = {1049--1065},
  publisher = {Association for Computational Linguistics},
  address = {Toronto, Canada},
  doi = {10.18653/v1/2023.findings-acl.67},
  urldate = {2026-01-07}
}

@article{weiEmergentAbilitiesLarge,
  title = {Emergent {{Abilities}} of {{Large Language Models}}},
  author = {Wei, Jason and Tay, Yi and Bommasani, Rishi and Raffel, Colin and Zoph, Barret and Borgeaud, Sebastian and Yogatama, Dani and Bosma, Maarten and Zhou, Denny and Metzler, Donald and Chi, Ed H and Hashimoto, Tatsunori and Vinyals, Oriol and Liang, Percy and Dean, Jeff and Fedus, William},
  journal = {Transactions on Machine Learning Research},
  year = 2022,
  langid = {english}
}

@inproceedings{weiChainofthoughtPromptingElicits2022,
  title = {Chain-of-Thought Prompting Elicits Reasoning in Large Language Models},
  booktitle = {Proceedings of the 36th {{International Conference}} on {{Neural Information Processing Systems}}},
  author = {Wei, Jason and Wang, Xuezhi and Schuurmans, Dale and Bosma, Maarten and Ichter, Brian and Xia, Fei and Chi, Ed H. and Le, Quoc V. and Zhou, Denny},
  year = 2022,
  month = nov,
  series = {{{NIPS}} '22},
  pages = {24824--24837},
  publisher = {Curran Associates Inc.},
  address = {Red Hook, NY, USA},
  urldate = {2025-03-06},
  isbn = {978-1-7138-7108-8}
}

@inproceedings{kojimaLargeLanguageModels2022,
  title = {Large Language Models Are Zero-Shot Reasoners},
  booktitle = {Proceedings of the 36th {{International Conference}} on {{Neural Information Processing Systems}}},
  author = {Kojima, Takeshi and Gu, Shixiang Shane and Reid, Machel and Matsuo, Yutaka and Iwasawa, Yusuke},
  year = 2022,
  month = nov,
  series = {{{NIPS}} '22},
  pages = {22199--22213},
  publisher = {Curran Associates Inc.},
  address = {Red Hook, NY, USA},
  urldate = {2025-03-06},
  isbn = {978-1-7138-7108-8}
}

@misc{wangSelfConsistencyImprovesChain2023,
  title = {Self-{{Consistency Improves Chain}} of {{Thought Reasoning}} in {{Language Models}}},
  author = {Wang, Xuezhi and Wei, Jason and Schuurmans, Dale and Le, Quoc and Chi, Ed and Narang, Sharan and Chowdhery, Aakanksha and Zhou, Denny},
  year = 2023,
  month = mar,
  number = {arXiv:2203.11171},
  eprint = {2203.11171},
  primaryclass = {cs},
  publisher = {arXiv},
  doi = {10.48550/arXiv.2203.11171},
  urldate = {2026-01-07},
  archiveprefix = {arXiv}
}

@inproceedings{trungReFTReasoningReinforced2024,
  title = {{{ReFT}}: {{Reasoning}} with {{Reinforced Fine-Tuning}}},
  shorttitle = {{{ReFT}}},
  booktitle = {Proceedings of the 62nd {{Annual Meeting}} of the {{Association}} for {{Computational Linguistics}} ({{Volume}} 1: {{Long Papers}})},
  author = {Trung, Luong and Zhang, Xinbo and Jie, Zhanming and Sun, Peng and Jin, Xiaoran and Li, Hang},
  editor = {Ku, Lun-Wei and Martins, Andre and Srikumar, Vivek},
  year = 2024,
  month = aug,
  pages = {7601--7614},
  publisher = {Association for Computational Linguistics},
  address = {Bangkok, Thailand},
  doi = {10.18653/v1/2024.acl-long.410},
  urldate = {2026-01-07}
}

@inproceedings{wangReinforcementLearningReasoning,
  title = {Reinforcement {{Learning}} for {{Reasoning}} in {{Large Language Models}} with {{One Training Example}}},
  booktitle = {39th {{Conference}} on {{Neural Information Processing Systems}} ({{NeurIPS}} 2025)},
  author = {Wang, Yiping and Yang, Qing and Zeng, Zhiyuan and Ren, Liliang and Liu, Liyuan and Peng, Baolin and Cheng, Hao and He, Xuehai and Wang, Kuan and Gao, Jianfeng and Chen, Weizhu and Wang, Shuohang and Du, Simon Shaolei and Shen, Yelong},
  langid = {english},
  year = 2025
}

@misc{cobbeTrainingVerifiersSolve2021a,
  title = {Training {{Verifiers}} to {{Solve Math Word Problems}}},
  author = {Cobbe, Karl and Kosaraju, Vineet and Bavarian, Mohammad and Chen, Mark and Jun, Heewoo and Kaiser, Lukasz and Plappert, Matthias and Tworek, Jerry and Hilton, Jacob and Nakano, Reiichiro and Hesse, Christopher and Schulman, John},
  year = 2021,
  month = nov,
  number = {arXiv:2110.14168},
  eprint = {2110.14168},
  primaryclass = {cs},
  publisher = {arXiv},
  doi = {10.48550/arXiv.2110.14168},
  urldate = {2026-01-07},
  archiveprefix = {arXiv}
}

@inproceedings{hendrycksMeasuringMathematicalProblem2021,
  title = {Measuring {{Mathematical Problem Solving With}} the {{MATH Dataset}}},
  booktitle = {35th {{Conference}} on {{Neural Information Processing Systems}} ({{NeurIPS}} 2021) {{Track}} on {{Datasets}} and {{Benchmarks}}},
  author = {Hendrycks, Dan and Burns, Collin and Kadavath, Saurav and Arora, Akul and Basart, Steven and Tang, Eric and Song, Dawn and Steinhardt, Jacob},
  year = 2021,
  month = nov,
  eprint = {2103.03874},
  primaryclass = {cs},
  publisher = {arXiv},
  doi = {10.48550/arXiv.2103.03874},
  urldate = {2026-01-07},
  archiveprefix = {arXiv}
}

@inproceedings{kazemiBIGBenchExtraHard2025a,
  title = {{{BIG-Bench Extra Hard}}},
  booktitle = {Proceedings of the 63rd {{Annual Meeting}} of the {{Association}} for {{Computational Linguistics}} ({{Volume}} 1: {{Long Papers}})},
  author = {Kazemi, Mehran and Fatemi, Bahare and Bansal, Hritik and Palowitch, John and Anastasiou, Chrysovalantis and Mehta, Sanket Vaibhav and Jain, Lalit K and Aglietti, Virginia and Jindal, Disha and Chen, Peter and Dikkala, Nishanth and Tyen, Gladys and Liu, Xin and Shalit, Uri and Chiappa, Silvia and Olszewska, Kate and Tay, Yi and Tran, Vinh Q. and Le, Quoc V and Firat, Orhan},
  editor = {Che, Wanxiang and Nabende, Joyce and Shutova, Ekaterina and Pilehvar, Mohammad Taher},
  year = 2025,
  month = jul,
  pages = {26473--26501},
  publisher = {Association for Computational Linguistics},
  address = {Vienna, Austria},
  doi = {10.18653/v1/2025.acl-long.1285},
  urldate = {2026-01-07},
  isbn = {979-8-89176-251-0}
}

@misc{cholletARCAGI2NewChallenge2025,
  title = {{{ARC-AGI-2}}: {{A New Challenge}} for {{Frontier AI Reasoning Systems}}},
  shorttitle = {{{ARC-AGI-2}}},
  author = {Chollet, Francois and Knoop, Mike and Kamradt, Gregory and Landers, Bryan and Pinkard, Henry},
  year = 2025,
  month = may,
  number = {arXiv:2505.11831},
  eprint = {2505.11831},
  primaryclass = {cs},
  publisher = {arXiv},
  doi = {10.48550/arXiv.2505.11831},
  urldate = {2026-01-07},
  archiveprefix = {arXiv}
}

@misc{roziereCodeLlamaOpen2024,
  title = {Code {{Llama}}: {{Open Foundation Models}} for {{Code}}},
  shorttitle = {Code {{Llama}}},
  author = {Rozi{\`e}re, Baptiste and Gehring, Jonas and Gloeckle, Fabian and Sootla, Sten and Gat, Itai and Tan, Xiaoqing Ellen and Adi, Yossi and Liu, Jingyu and Sauvestre, Romain and Remez, Tal and Rapin, J{\'e}r{\'e}my and Kozhevnikov, Artyom and Evtimov, Ivan and Bitton, Joanna and Bhatt, Manish and Ferrer, Cristian Canton and Grattafiori, Aaron and Xiong, Wenhan and D{\'e}fossez, Alexandre and Copet, Jade and Azhar, Faisal and Touvron, Hugo and Martin, Louis and Usunier, Nicolas and Scialom, Thomas and Synnaeve, Gabriel},
  year = 2024,
  month = jan,
  number = {arXiv:2308.12950},
  eprint = {2308.12950},
  primaryclass = {cs},
  publisher = {arXiv},
  doi = {10.48550/arXiv.2308.12950},
  urldate = {2026-01-09},
  archiveprefix = {arXiv}
}

@misc{yangCodeThinkThink2025,
  title = {Code to {{Think}}, {{Think}} to {{Code}}: {{A Survey}} on {{Code-Enhanced Reasoning}} and {{Reasoning-Driven Code Intelligence}} in {{LLMs}}},
  shorttitle = {Code to {{Think}}, {{Think}} to {{Code}}},
  author = {Yang, Dayu and Liu, Tianyang and Zhang, Daoan and Simoulin, Antoine and Liu, Xiaoyi and Cao, Yuwei and Teng, Zhaopu and Qian, Xin and Yang, Grey and Luo, Jiebo and McAuley, Julian},
  year = 2025,
  month = feb,
  number = {arXiv:2502.19411},
  eprint = {2502.19411},
  primaryclass = {cs},
  publisher = {arXiv},
  doi = {10.48550/arXiv.2502.19411},
  urldate = {2025-10-22},
  archiveprefix = {arXiv}
}

@misc{aryabumiCodeNotCode2024,
  title = {To {{Code}}, or {{Not To Code}}? {{Exploring Impact}} of {{Code}} in {{Pre-training}}},
  shorttitle = {To {{Code}}, or {{Not To Code}}?},
  author = {Aryabumi, Viraat and Su, Yixuan and Ma, Raymond and Morisot, Adrien and Zhang, Ivan and Locatelli, Acyr and Fadaee, Marzieh and {\"U}st{\"u}n, Ahmet and Hooker, Sara},
  year = 2024,
  month = aug,
  number = {arXiv:2408.10914},
  eprint = {2408.10914},
  primaryclass = {cs},
  publisher = {arXiv},
  doi = {10.48550/arXiv.2408.10914},
  urldate = {2025-10-16},
  archiveprefix = {arXiv}
}

@misc{linScalingCodeAssistedChainofThoughts2025,
  title = {Scaling {{Code-Assisted Chain-of-Thoughts}} and {{Instructions}} for {{Model Reasoning}}},
  author = {Lin, Honglin and Pei, Qizhi and Gao, Xin and Pan, Zhuoshi and Li, Yu and Li, Juntao and He, Conghui and Wu, Lijun},
  year = 2025,
  month = oct,
  number = {arXiv:2510.04081},
  eprint = {2510.04081},
  primaryclass = {cs},
  publisher = {arXiv},
  doi = {10.48550/arXiv.2510.04081},
  urldate = {2025-12-02},
  archiveprefix = {arXiv}
}

@misc{maWhichTrainingStage2023,
  title = {At {{Which Training Stage Does Code Data Help LLMs Reasoning}}?},
  author = {Ma, Yingwei and Liu, Yue and Yu, Yue and Zhang, Yuanliang and Jiang, Yu and Wang, Changjian and Li, Shanshan},
  year = 2023,
  month = sep,
  number = {arXiv:2309.16298},
  eprint = {2309.16298},
  primaryclass = {cs},
  publisher = {arXiv},
  doi = {10.48550/arXiv.2309.16298},
  urldate = {2025-10-16},
  archiveprefix = {arXiv}
}

@misc{grattafioriLlama3Herd2024,
  title = {The {{Llama}} 3 {{Herd}} of {{Models}}},
  author = {Grattafiori, Aaron and Dubey, Abhimanyu and Jauhri, Abhinav and others},
  year = 2024,
  month = nov,
  number = {arXiv:2407.21783},
  eprint = {2407.21783},
  primaryclass = {cs},
  publisher = {arXiv},
  doi = {10.48550/arXiv.2407.21783},
  urldate = {2024-12-17},
  archiveprefix = {arXiv}
}

@misc{qwenQwen25TechnicalReport2025,
  title = {Qwen2.5 {{Technical Report}}},
  author = {Qwen and Yang, An and Yang, Baosong and Zhang, Beichen and Hui, Binyuan and Zheng, Bo and Yu, Bowen and Li, Chengyuan and Liu, Dayiheng and Huang, Fei and Wei, Haoran and Lin, Huan and Yang, Jian and Tu, Jianhong and Zhang, Jianwei and Yang, Jianxin and Yang, Jiaxi and Zhou, Jingren and Lin, Junyang and Dang, Kai and Lu, Keming and Bao, Keqin and Yang, Kexin and Yu, Le and Li, Mei and Xue, Mingfeng and Zhang, Pei and Zhu, Qin and Men, Rui and Lin, Runji and Li, Tianhao and Tang, Tianyi and Xia, Tingyu and Ren, Xingzhang and Ren, Xuancheng and Fan, Yang and Su, Yang and Zhang, Yichang and Wan, Yu and Liu, Yuqiong and Cui, Zeyu and Zhang, Zhenru and Qiu, Zihan},
  year = 2025,
  month = jan,
  number = {arXiv:2412.15115},
  eprint = {2412.15115},
  primaryclass = {cs},
  publisher = {arXiv},
  doi = {10.48550/arXiv.2412.15115},
  urldate = {2025-06-05},
  archiveprefix = {arXiv}
}

@inproceedings{huLoRALowRankAdaptation2021a,
  title = {{{LoRA}}: {{Low-Rank Adaptation}} of {{Large Language Models}}},
  shorttitle = {{{LoRA}}},
  booktitle = {International {{Conference}} on {{Learning Representations}}},
  author = {Hu, Edward J. and Shen, Yelong and Wallis, Phillip and {Allen-Zhu}, Zeyuan and Li, Yuanzhi and Wang, Shean and Wang, Lu and Chen, Weizhu},
  year = 2021,
  month = oct,
  urldate = {2025-11-17},
  langid = {english}
}

@inproceedings{lightmanLetsVerifyStep2023,
  title = {Let's {{Verify Step}} by {{Step}}},
  booktitle = {12th {{International Conference}} on {{Learning Representations}} ({{ICLR24}})},
  author = {Lightman, Hunter and Kosaraju, Vineet and Burda, Yura and Edwards, Harri and Baker, Bowen and Lee, Teddy and Leike, Jan and Schulman, John and Sutskever, Ilya and Cobbe, Karl},
  year = 2023,
  month = may,
  eprint = {2305.20050},
  primaryclass = {cs},
  urldate = {2026-01-12},
  archiveprefix = {arXiv}
}

@misc{phanHumanitysLastExam2025,
  title = {Humanity's {{Last Exam}}},
  author = {Phan, Long and Gatti, Alice and Han, Ziwen and others},
  year = 2025,
  month = sep,
  number = {arXiv:2501.14249},
  eprint = {2501.14249},
  primaryclass = {cs},
  publisher = {arXiv},
  doi = {10.48550/arXiv.2501.14249},
  urldate = {2026-01-12},
  archiveprefix = {arXiv}
}

@misc{reinGPQAGraduateLevelGoogleProof2023,
  title = {{{GPQA}}: {{A Graduate-Level Google-Proof Q}}\&{{A Benchmark}}},
  shorttitle = {{{GPQA}}},
  author = {Rein, David and Hou, Betty Li and Stickland, Asa Cooper and Petty, Jackson and Pang, Richard Yuanzhe and Dirani, Julien and Michael, Julian and Bowman, Samuel R.},
  year = 2023,
  month = nov,
  number = {arXiv:2311.12022},
  eprint = {2311.12022},
  primaryclass = {cs},
  publisher = {arXiv},
  doi = {10.48550/arXiv.2311.12022},
  urldate = {2026-01-12},
  archiveprefix = {arXiv}
}

@misc{yuanHowWellLarge2023,
  title = {How Well Do {{Large Language Models}} Perform in {{Arithmetic}} Tasks?},
  author = {Yuan, Zheng and Yuan, Hongyi and Tan, Chuanqi and Wang, Wei and Huang, Songfang},
  year = 2023,
  month = mar,
  number = {arXiv:2304.02015},
  eprint = {2304.02015},
  primaryclass = {cs},
  publisher = {arXiv},
  doi = {10.48550/arXiv.2304.02015},
  urldate = {2026-01-12},
  archiveprefix = {arXiv}
}

@misc{meta-llamaMetallamaLlama323BInstructHugging2024,
  title = {Meta-Llama/{{Llama-3}}.2-{{3B-Instruct}} {$\cdot$} {{Hugging Face}}},
  author = {{meta-llama}},
  year = 2024,
  month = dec,
  urldate = {2026-01-12},
  howpublished = {https://huggingface.co/meta-llama/Llama-3.2-3B-Instruct}
}

@misc{qwenQwenQwen253BInstructHugging2025,
  title = {Qwen/{{Qwen2}}.5-{{3B-Instruct}} {$\cdot$} {{Hugging Face}}},
  author = {Qwen},
  year = 2025,
  month = dec,
  urldate = {2026-01-12},
  howpublished = {https://huggingface.co/Qwen/Qwen2.5-3B-Instruct}
}

@misc{qwenQwenQwen257BInstructHugging2025,
  title = {Qwen/{{Qwen2}}.5-{{7B-Instruct}} {$\cdot$} {{Hugging Face}}},
  author = {Qwen},
  year = 2025,
  month = dec,
  urldate = {2026-01-12},
  howpublished = {https://huggingface.co/Qwen/Qwen2.5-7B-Instruct}
}

@misc{qwenQwenQwen2514BInstructHugging2025,
  title = {Qwen/{{Qwen2}}.5-{{14B-Instruct}} {$\cdot$} {{Hugging Face}}},
  author = {Qwen},
  year = 2025,
  month = dec,
  urldate = {2026-01-19},
  howpublished = {https://huggingface.co/Qwen/Qwen2.5-14B-Instruct}
}

@inproceedings{fengRevealingMysteryChain2023,
  title = {Towards Revealing the Mystery behind Chain of Thought: A Theoretical Perspective},
  shorttitle = {Towards Revealing the Mystery behind Chain of Thought},
  booktitle = {Proceedings of the 37th {{International Conference}} on {{Neural Information Processing Systems}}},
  author = {Feng, Guhao and Zhang, Bohang and Gu, Yuntian and Ye, Haotian and He, Di and Wang, Liwei},
  year = 2023,
  month = dec,
  series = {{{NIPS}} '23},
  pages = {70757--70798},
  publisher = {Curran Associates Inc.},
  address = {Red Hook, NY, USA},
  urldate = {2026-01-19}
}

@misc{chenProgramThoughtsPrompting2023a,
  title = {Program of {{Thoughts Prompting}}: {{Disentangling Computation}} from {{Reasoning}} for {{Numerical Reasoning Tasks}}},
  shorttitle = {Program of {{Thoughts Prompting}}},
  author = {Chen, Wenhu and Ma, Xueguang and Wang, Xinyi and Cohen, William W.},
  year = 2023,
  month = oct,
  number = {arXiv:2211.12588},
  eprint = {2211.12588},
  primaryclass = {cs},
  publisher = {arXiv},
  doi = {10.48550/arXiv.2211.12588},
  urldate = {2026-01-19},
  archiveprefix = {arXiv}
}

@article{gillCyclomaticComplexityDensity1991,
  title = {Cyclomatic {{Complexity Density}} and {{Software Maintenance Productivity}}},
  author = {Gill, Geoffrey K. and Kemerer, Chris F.},
  year = 1991,
  journal = {IEEE Trans. Software Eng.},
  volume = {17},
  number = {12},
  pages = {1284--1288},
  doi = {10.1109/32.106988}
}

@misc{jiangMistral7B2023,
  title = {Mistral {{7B}}},
  author = {Jiang, Albert Q. and Sablayrolles, Alexandre and Mensch, Arthur and Bamford, Chris and Chaplot, Devendra Singh and de las Casas, Diego and Bressand, Florian and Lengyel, Gianna and Lample, Guillaume and Saulnier, Lucile and Lavaud, L{\'e}lio Renard and Lachaux, Marie-Anne and Stock, Pierre and Scao, Teven Le and Lavril, Thibaut and Wang, Thomas and Lacroix, Timoth{\'e}e and Sayed, William El},
  year = 2023,
  month = oct,
  number = {arXiv:2310.06825},
  eprint = {2310.06825},
  primaryclass = {cs},
  publisher = {arXiv},
  doi = {10.48550/arXiv.2310.06825},
  urldate = {2024-12-17},
  archiveprefix = {arXiv}
}

@misc{mistralaiMistralaiMistral7BInstructv03Hugging,
  title = {Mistralai/{{Mistral-7B-Instruct-v0}}.3 {$\cdot$} {{Hugging Face}}},
  author = {{mistralai}},
  urldate = {2026-01-20},
  year = 2025,
  howpublished = {https://huggingface.co/mistralai/Mistral-7B-Instruct-v0.3}
}

@article{liuDatasetsLargeLanguage2025,
  title = {Datasets for Large Language Models: A Comprehensive Survey},
  shorttitle = {Datasets for Large Language Models},
  author = {Liu, Yang and Cao, Jiahuan and Liu, Chongyu and Ding, Kai and Jin, Lianwen},
  year = 2025,
  month = oct,
  journal = {Artificial Intelligence Review},
  volume = {58},
  number = {12},
  pages = {403},
  issn = {1573-7462},
  doi = {10.1007/s10462-025-11403-7},
  urldate = {2026-01-26},
  langid = {english}
}

@inproceedings{abedIncreasingLLMCoding2025,
  title = {Increasing {{LLM Coding Capabilities}} through {{Diverse Synthetic Coding Tasks}}},
  booktitle = {{{NeurIPS}} 2025 {{Fourth Workshop}} on {{Deep Learning}} for {{Code}}},
  author = {Abed, Amal and Lukic, Ivan and Franke, J{\"o}rg K. H. and Hutter, Frank},
  year = 2025,
  month = nov,
  urldate = {2026-01-26},
  langid = {english}
}

@inproceedings{ainslieGQATrainingGeneralized2023,
  title = {{{GQA}}: {{Training Generalized Multi-Query Transformer Models}} from {{Multi-Head Checkpoints}}},
  shorttitle = {{{GQA}}},
  booktitle = {Proceedings of the 2023 {{Conference}} on {{Empirical Methods}} in {{Natural Language Processing}}},
  author = {Ainslie, Joshua and {Lee-Thorp}, James and {de Jong}, Michiel and Zemlyanskiy, Yury and Lebron, Federico and Sanghai, Sumit},
  editor = {Bouamor, Houda and Pino, Juan and Bali, Kalika},
  year = 2023,
  month = dec,
  pages = {4895--4901},
  publisher = {Association for Computational Linguistics},
  address = {Singapore},
  doi = {10.18653/v1/2023.emnlp-main.298},
  urldate = {2026-01-27}
}

@incollection{dodgeSpearmanRankCorrelation2008,
  title = {Spearman {{Rank Correlation Coefficient}}},
  booktitle = {The {{Concise Encyclopedia}} of {{Statistics}}},
  author = {Dodge, Yadolah},
  year = 2008,
  pages = {502--505},
  publisher = {Springer, New York, NY},
  doi = {10.1007/978-0-387-32833-1_379},
  urldate = {2026-01-27},
  isbn = {978-0-387-32833-1},
  langid = {english}
}

@inproceedings{longprePretrainersGuideTraining2024,
  title = {A {{Pretrainer}}'s {{Guide}} to {{Training Data}}: {{Measuring}} the {{Effects}} of {{Data Age}}, {{Domain Coverage}}, {{Quality}}, \& {{Toxicity}}},
  shorttitle = {A {{Pretrainer}}'s {{Guide}} to {{Training Data}}},
  booktitle = {Proceedings of the 2024 {{Conference}} of the {{North American Chapter}} of the {{Association}} for {{Computational Linguistics}}: {{Human Language Technologies}} ({{Volume}} 1: {{Long Papers}})},
  author = {Longpre, Shayne and Yauney, Gregory and Reif, Emily and Lee, Katherine and Roberts, Adam and Zoph, Barret and Zhou, Denny and Wei, Jason and Robinson, Kevin and Mimno, David and Ippolito, Daphne},
  editor = {Duh, Kevin and Gomez, Helena and Bethard, Steven},
  year = 2024,
  month = jun,
  pages = {3245--3276},
  publisher = {Association for Computational Linguistics},
  address = {Mexico City, Mexico},
  doi = {10.18653/v1/2024.naacl-long.179},
  urldate = {2026-01-27}
}

@inproceedings{chenMasteringCraftData2025,
  title = {Mastering the {{Craft}} of {{Data Synthesis}} for {{CodeLLMs}}},
  booktitle = {Proceedings of the 2025 {{Conference}} of the {{Nations}} of the {{Americas Chapter}} of the {{Association}} for {{Computational Linguistics}}: {{Human Language Technologies}} ({{Volume}} 1: {{Long Papers}})},
  author = {Chen, Meng and Arthur, Philip and Feng, Qianyu and Hoang, Cong Duy Vu and Hong, Yu-Heng and Moghaddam, Mahdi Kazemi and Nezami, Omid and Nguyen, Duc Thien and Tangari, Gioacchino and Vu, Duy and Vu, Thanh and Johnson, Mark and Kenthapadi, Krishnaram and Dharmasiri, Don and Duong, Long and Li, Yuan-Fang},
  editor = {Chiruzzo, Luis and Ritter, Alan and Wang, Lu},
  year = 2025,
  month = apr,
  pages = {12484--12500},
  publisher = {Association for Computational Linguistics},
  address = {Albuquerque, New Mexico},
  doi = {10.18653/v1/2025.naacl-long.620},
  urldate = {2026-01-27},
  isbn = {979-8-89176-189-6}
}

@misc{luoEmpiricalStudyCatastrophic2025,
  title = {An {{Empirical Study}} of {{Catastrophic Forgetting}} in {{Large Language Models During Continual Fine-tuning}}},
  author = {Luo, Yun and Yang, Zhen and Meng, Fandong and Li, Yafu and Zhou, Jie and Zhang, Yue},
  year = 2025,
  month = jan,
  number = {arXiv:2308.08747},
  eprint = {2308.08747},
  primaryclass = {cs},
  publisher = {arXiv},
  doi = {10.48550/arXiv.2308.08747},
  urldate = {2026-01-28},
  archiveprefix = {arXiv}
}
\bibliographystyle{icml2026}


\newpage
\appendix
\onecolumn


\section{Complexity Metrics Implementation Details.}
\label{app:detail_complexity}

To characterise the complexity of the code used during fine-tuning, we compute two complementary static metrics over all samples: cyclomatic complexity (\cc) and logical lines of code (\lloc).
This appendix reports the implementation details of their calculation.

\subsection{Complexity Metric Calculation}

We calculate our complexity metrics using established open-source static analysis tools, here we provide details on our implementation of these tools.
For both datasets, we first ensure that we have the raw solution code extracted from the natural language response: for \codenet, the code is already provided separately; for \instruct, the natural language responses are given in \texttt{Markdown} format, therefore we can use \texttt{regex} matching to extract code blocks and programming languages by searching for a triple backtick followed by a programming language name~\citep{ExtendedSyntaxMarkdown2025}. 
We then process the code for each language (\python, \javascript, \java) separately:

\begin{itemize}

    \item 
        For \textbf{\python}, we install and use the \texttt{Radon} library\footnote{Radon: \href{https://pypi.org/project/radon/}{https://pypi.org/project/radon/}}.
        We use the \texttt{radon.complexity.cc\_visit} function to calculate \cc;
        and the \texttt{radon.raw.analyze} function to calculate \lloc.
        
    \item
        For \textbf{\javascript}, we use the \texttt{complexity-report}\footnote{complexity-report: \href{https://www.npmjs.com/package/complexity-report}{https://www.npmjs.com/package/complexity-report}} command-line tool, to run the \texttt{escomplex} library\footnote{escomplex: \href{https://www.npmjs.com/package/escomplex}{https://www.npmjs.com/package/escomplex}} against our code.
        We use the \texttt{cr} command with the \texttt{--format json} option, to produce a report that contains both \cc\ and \lloc.
        
    \item 
        For \textbf{\java}, we use the \texttt{PMD} static code analyzer\footnote{PMD: \href{https://pmd.github.io/}{https://pmd.github.io/}}, to be run from the command line.
        We use the \texttt{pmd} command with the \texttt{-f json} option, and passing in a ruleset that requests \texttt{CyclomaticComplexity} and \texttt{NcssCount}, to produce a report that contains both \cc\ and \lloc.

\end{itemize}

\subsection{Complexity Metric Aggregation}

Cyclomatic complexity is defined at the level of individual functions; following common practice in software metrics, we aggregate function-level values to the solution level by taking the maximum over all functions included.
This reflects the intuition that structural complexity is often dominated by the most complex execution path~\cite{gillCyclomaticComplexityDensity1991}.
By contrast, \lloc\ acts as an additive size-based measure and therefore does not require special handling--we simply sum all logical lines of code contained in the solution.


\section{CodeNet Augmentation Prompts.}
\label{app:augment}

To convert Project CodeNet~\cite{puriCodeNetLargeScaleAI2021a} into an instruction--response format suitable for supervised fine-tuning, we apply an LLM-based augmentation step using a fixed system prompt and two task-specific user prompts.
All prompts are applied to each CodeNet problem--solution pair to ensure consistency across augmented samples.
We generated responses to the prompts using the \texttt{gpt-5-mini-2025-08-07} model~\cite{openaiGPT5MiniAPI2025} via the OpenAI API\footnote{OpenAI API: \href{https://openai.com/api/}{https://openai.com/api/}}, with the default API parameters (\texttt{reasoning.effort = medium} and \texttt{text.verbosity = medium}).

\paragraph{System prompt.}
All augmentation calls use the same system prompt, which constrains the model to return only the requested content without additional commentary:

\begin{quote}
\small
\texttt{You are a helpful assistant that will assist in creating a new code-based benchmark dataset. When responding, you only provide exactly what is requested, with no additional text.}
\end{quote}

\paragraph{Instruction-template prompt.}
To convert HTML problem statements into natural-language instructions, we prompt the model to produce a concise, language-agnostic instruction that preserves the original task specification exactly and includes a \texttt{<language>} placeholder to be substituted later.
A single instruction template is generated per CodeNet record.

\begin{quote}
\small
\texttt{I am augmenting the Project CodeNet (by IBM) dataset, converting it into an instruction / response dataset that can be used for supervised finetuning, and I need assistance.\\
The current problem statements are provided in HTML, and I need you to convert them into a natrual language prompt instruction that I can use to ask models to generate code.\\
The instruction must be programming language agnostic, but you must provide a <language> token in the instruction, that I can replace with the programming language that must be used.\\
It is vital that the requested specifications are exactly the same as the original.\\
Only provide the instruction exactly as it should be used in the dataset.
Here is the original HTML problem statement:\\
\textbf{\{html\}}
}
\end{quote}

\paragraph{Response-template prompt.}
To standardise model outputs, we prompt the model to generate a natural-sounding response template that wraps the solution code in surrounding explanatory text.
The template is language-agnostic and contains two placeholders: \texttt{<language>} for the programming language and \texttt{<code>} for the solution code, which is inserted verbatim during dataset construction.
Three different response templates are generated per CodeNet record, so that each programming language has a unique response template.

\begin{quote}
\small
\texttt{I am augmenting the Project CodeNet (by IBM) dataset, converting it into an instruction / response dataset that can be used for supervised finetuning, and I need assistance.\\
The solutions are currently provided as just raw code, and I need your help to turn them into readable and useful model responses that can be used for training.\\
I need you to provide a template for a response that would read naturally to a user, you should add the surrounding text that LLMs typically provide, reading as if it is a real response from an LLM solving the task. Do not include specifics of the code, approach or algorithm though - you have not seen the code yet, so it might not be accurate.\\
The response should be language agnostic (do not use those words though), but must contain a <language> token, that I can replace with the programming language that is used.\\
You must also provide a <code> token that I will replace with the code block of the response, there is no need for a corresponding </code>.\\
The <code> token must be surrounded by newlines, but I will handle correctly having the code itself contained within triple backticks (as per markdown).\\
Only provide the response template exactly as it should be used in the dataset.\\
Here is the instruction:\\
\textbf{\{instruction\}}
}
\end{quote}


\section{Dataset Statistics.}
\label{app:datasets}

This appendix reports additional statistics for the datasets used in our fine-tuning experiments.
We create a solution-driven complexity dataset (\codenet) and a problem-driven complexity dataset (\instruct), each of which has twelve splits: five different complexity levels (\minspl, \lowspl, \midspl, \highspl, \maxspl) and a control (\ctrlspl) for each complexity metric (\cc and \lloc).

\subsection{Language Statistics}
All splits in the \codenet\ dataset contain 2,919 \python\ samples, 1,890 \javascript\ samples and 3,278 \java\ samples.
All splits in the \instruct\ dataset contain 3,688 \python\ samples, 2,789 \javascript\ samples and 1,610 \java\ samples.

\subsection{Metric Statistics}
Table~\ref{tab:complexity-stats} reports the mean \cc\ and \lloc\ values for each split across both datasets.
Reporting both metrics is important because \cc\ and \lloc\ are correlated in practice; including \lloc\ helps control for confounding effects of code length when assessing structural complexity.

\begin{table*}[h]

    \centering
    \caption{
        \textit{\textbf{Structural complexity statistics.}}
        We report mean cyclomatic complexity (\cc) and logical lines of code (\lloc) for each split of the \codenet\ (solution-driven complexity) and \instruct\ (problem-driven complexity) datasets.
    }
    \label{tab:complexity-stats}
    
    \small
    \setlength{\tabcolsep}{8pt}

    \begin{tabular}{lcccccccc}
        \toprule
        
        & \multicolumn{4}{c}{\thead{\codenet: solution-driven complexity}} & \multicolumn{4}{c}{\thead{\instruct: problem-driven complexity}} \\
        \cmidrule(lr){2-5} \cmidrule(lr){6-9}
        \textbf{Split} 
        & \multicolumn{2}{c}{\thead{\cc\ dataset split}} & \multicolumn{2}{c}{\thead{\lloc\ dataset split}} 
        & \multicolumn{2}{c}{\thead{\cc\ dataset split}} & \multicolumn{2}{c}{\thead{\lloc\ dataset split}} \\
        \cmidrule(lr){2-3} \cmidrule(lr){4-5} \cmidrule(lr){6-7} \cmidrule(lr){8-9}
        & Avg. \cc & Avg. \lloc & Avg. \cc & Avg. \lloc & Avg. \cc & Avg. \lloc & Avg. \cc & Avg. \lloc \\
        
        \midrule
        \minspl\  & 7.79 & 37.67 & 8.83 & 32.67 & 0.63 & 8.35 & 0.96 & 3.43 \\
        \lowspl\      & 9.91 & 44.65 & 10.66 & 40.75 & 0.69 & 8.26 & 1.30 & 4.74 \\
        \midspl\   & 13.99 & 58.81 & 14.32 & 57.69 & 2.00 & 10.76 & 2.62 & 9.65 \\
        \highspl\     & 20.63 & 82.50 & 19.95 & 84.82 & 3.89 & 15.25 & 4.01 & 16.06 \\
        \maxspl\  & 43.03 & 169.67 & 40.63 & 180.17 & 11.12 & 29.83 & 7.58 & 43.94 \\
        \ctrlspl\  & 18.84 & 76.14 & 19.21 & 81.83 & 3.78 & 15.02 & 3.78 & 15.81 \\
        \bottomrule
    \end{tabular}

\end{table*}


\section{Model Evaluation Details.}
\label{app:models}

\subsection{Model Details}

Table~\ref{tab:models} reports the full details of all models used in our experiments.
We document parameter scale, context window, provider, and knowledge cut-off to make transparent the architectural and training differences across models and to facilitate reproducibility of our results.
All models are publicly available through Hugging Face\footnote{Hugging Face: \href{https://huggingface.co/}{https://huggingface.co/}}, and we use only officially released instruct variants to ensure consistent evaluation behaviour across tasks.
All models employ grouped-query attention~\cite{ainslieGQATrainingGeneralized2023}; therefore attention heads are reported as the number of query heads and key--value heads (Q/KV).

\begin{table*}[h]
    \centering
    
    \caption{
        \textit{\textbf{Full details for all models evaluated in our experiments.}}
        We report model size, context length, and architectural characteristics for each model.
    }

    \label{tab:models}
    
    \begin{adjustbox}{width=\textwidth}
        
        \begin{tabular}{llrrcccl}
        \toprule
        \thead{Model Name} &
        \thead{Full Model Version} &
        \thead{Parameters} &
        \thead{Context\\Length} &
        \thead{Attention\\Heads (Q/KV)} &
        \thead{Provider} &
        \thead{Knowledge\\Cutoff} &
        \thead{Source} \\
        
        \midrule
        
        \qwmini &
        \texttt{Qwen2.5-3B-Instruct} &
        3.09B &
        32K &
        16 / 2 &
        Alibaba &
        Dec.\ 2023 &
        \href{https://huggingface.co/Qwen/Qwen2.5-3B-Instruct}{Hugging Face}~\cite{qwenQwenQwen253BInstructHugging2025} \\
        
        \qwsmall &
        \texttt{Qwen2.5-7B-Instruct} &
        7.61B &
        32K &
        28 / 4 &
        Alibaba &
        Dec.\ 2023 &
        \href{https://huggingface.co/Qwen/Qwen2.5-7B-Instruct}{Hugging Face}~\cite{qwenQwenQwen257BInstructHugging2025} \\
        
        \qwmedium &
        \texttt{Qwen2.5-14B-Instruct} &
        14.70B &
        128K &
        40 / 8 &
        Alibaba &
        Dec.\ 2023 &
        \href{https://huggingface.co/Qwen/Qwen2.5-14B-Instruct}{Hugging Face}~\cite{qwenQwenQwen2514BInstructHugging2025} \\
        
        \llmini &
        \texttt{Llama-3.2-3B-Instruct} &
        3.21B &
        128K &
        24 / 8 &
        Meta &
        Dec.\ 2023 &
        \href{https://huggingface.co/meta-llama/Llama-3.2-3B-Instruct}{Hugging Face}~\cite{meta-llamaMetallamaLlama323BInstructHugging2024} \\
        
        \llsmall &
        \texttt{Llama-3.1-8B-Instruct} &
        8.00B &
        128K &
        32 / 8 &
        Meta &
        Dec.\ 2023 &
        \href{https://huggingface.co/meta-llama/Llama-3.1-8B-Instruct}{Hugging Face}~\cite{meta-llamaMetallamaLlama323BInstructHugging2024} \\
        
        \mistral &
        \texttt{Mistral-7B-Instruct-v0.3} &
        7.00B &
        32K &
        32 / 8 &
        Mistral &
        May 2024 &
        \href{https://huggingface.co/mistralai/Mistral-7B-Instruct-v0.3}{Hugging Face}~\cite{mistralaiMistralaiMistral7BInstructv03Hugging} \\
        
        \bottomrule
        \end{tabular}
    
    \end{adjustbox}

\end{table*}

\subsection{Training configuration}

We fine-tune all models using Low-Rank Adaptation (LoRA)~\cite{huLoRALowRankAdaptation2021a}.
While we generally follow standard practices, we specifically set the LoRA rank $r=16$, alpha $\alpha=16$, and dropout to $0$.
We apply LoRA adapters to all linear modules, including \texttt{q\_proj}, \texttt{k\_proj}, \texttt{v\_proj}, \texttt{o\_proj}, \texttt{gate\_proj}, \texttt{up\_proj}, and \texttt{down\_proj}.
We use the AdamW optimizer (\texttt{adamw\_torch}) with a learning rate of $2 \times 10^{-5}$, a cosine learning rate scheduler, and a warm-up ratio of $0.1$.
Training is conducted for 2 epochs with a per-device batch size of 4 and 4 gradient accumulation steps, resulting in an effective batch size of 16.
All models are trained with a maximum sequence length of 32,768 tokens using \texttt{bfloat16} precision.

\subsection{Evaluation configuration}

We maximize reproducibility by using a greedy decoding strategy with a temperature of $0.0$, top-p of $1.0$, and a maximum generation length of 16,384 tokens.
To standardize input formats across models, we append the following suffix to every query: ``\texttt{\textbackslash n Please reason step by step, and put your final answer within \textbackslash boxed\{\}.}''.
We extract the final answer by parsing the content within the \texttt{\textbackslash boxed\{\}} delimiters or other standard indicators (e.g., \texttt{<answer>}, \texttt{\#\#\#}). To ensure robust evaluation, we utilize the \texttt{math\_verify} library\footnote{Math-Verify: \href{https://github.com/huggingface/math-verify}{https://github.com/huggingface/math-verify}} to match the extracted answers against the ground truth.


\section{Comprehensive Results}
\label{app:results}

\Cref{tab:app-results} reports the complete set of per-benchmark results for all models and training configurations considered in this study.
For each benchmark, we include accuracies obtained after fine-tuning on complexity-controlled code datasets under both the solution-driven (\codenet) and problem-driven (\instruct) settings, split by cyclomatic complexity (CC) and logical lines of code (LLOC), alongside the corresponding control (\ctrlspl) datasets 
and the natural language (NL) baseline.
This table is provided for completeness and to support detailed inspection of individual model–benchmark behaviours.

\begin{table*}[h]
\centering

\caption{
    \textbf{\textit{Full per-benchmark results for all models.}}
    Average accuracy (\%) on each reasoning benchmark following fine-tuning on complexity-controlled code subsets.
    We report results for fine-tuning on solution-driven (\codenet) and problem-driven (\instruct) datasets, split by cyclomatic complexity (CC) and logical lines of code (LLOC), together with control splits and the natural language (NL) baseline, where the model is fine-tuned on a strictly non-code dataset.
}

\label{tab:app-results}
\resizebox{\textwidth}{!}{%
\begin{tabular}{llc cccccc cccccc cccccc cccccc}
\toprule
\textbf{} & \textbf{Model}
& \textbf{Baseline}
& \multicolumn{6}{c}{\textbf{CodeNet: CC splits}}
& \multicolumn{6}{c}{\textbf{CodeNet: LLOC splits}}
& \multicolumn{6}{c}{\textbf{Instruct: CC splits}}
& \multicolumn{6}{c}{\textbf{Instruct: LLOC splits}} \\
\cmidrule(lr){3-3}
\cmidrule(lr){4-9}
\cmidrule(lr){10-15}
\cmidrule(lr){16-21}
\cmidrule(lr){22-27}
& & \textbf{NL}
& \minspl\ & \lowspl\ & \midspl\ & \highspl\ & \maxspl\ & \ctrlspl\
& \minspl\ & \lowspl\ & \midspl\ & \highspl\ & \maxspl\ & \ctrlspl\
& \minspl\ & \lowspl\ & \midspl\ & \highspl\ & \maxspl\ & \ctrlspl\
& \minspl\ & \lowspl\ & \midspl\ & \highspl\ & \maxspl\ & \ctrlspl\ \\
\midrule
\addlinespace

\multirow{6}{*}{\rotatebox{90}{\textbf{\bbeh}}}
& \qwmini    & 6.1 & 8.9 & 11.5 & 9.6 & 7.8 & 8.5 & 9.6 & 10.2 & 10.0 & 8.7 & 10.4 & 10.0 & 10.4 & 8.3 & 7.4 & 6.7 & 7.0 & 7.2 & 7.2 & 7.8 & 8.9 & 7.4 & 6.5 & 7.6 & 6.7 \\
& \qwsmall   & 9.3 & 10.4 & 12.2 & 9.8 & 10.0 & 11.5 & 9.6 & 12.0 & 11.5 & 10.4 & 10.9 & 10.7 & 10.7 & 10.7 & 9.1 & 6.5 & 8.7 & 9.6 & 10.0 & 7.8 & 7.0 & 8.0 & 11.5 & 9.3 & 8.7 \\
& \qwmedium  & 12.8 & 12.0 & 12.6 & 14.6 & 13.7 & 14.8 & 13.0 & 12.4 & 14.1 & 12.6 & 15.7 & 13.9 & 13.5 & 12.4 & 12.6 & 12.6 & 13.3 & 13.3 & 12.6 & 12.2 & 14.3 & 13.5 & 13.5 & 13.9 & 14.1 \\
& \llmini    & 0.4 & 3.5 & 2.6 & 2.8 & 3.0 & 3.9 & 3.0 & 3.0 & 2.0 & 2.6 & 2.4 & 4.3 & 4.1 & 2.2 & 2.8 & 2.4 & 3.3 & 3.0 & 2.8 & 3.3 & 1.7 & 3.0 & 2.2 & 1.5 & 3.0 \\
& \llsmall   & 3.0 & 7.4 & 6.3 & 6.7 & 8.5 & 5.9 & 7.0 & 7.2 & 7.4 & 7.8 & 5.9 & 6.3 & 6.5 & 7.2 & 7.8 & 6.7 & 8.3 & 6.3 & 8.7 & 5.7 & 7.8 & 5.7 & 7.4 & 6.7 & 7.2 \\
& \mistral   & 5.2 & 5.7 & 5.2 & 4.8 & 4.6 & 6.3 & 6.3 & 5.9 & 5.0 & 5.7 & 4.6 & 5.2 & 5.2 & 5.2 & 4.6 & 5.0 & 3.9 & 4.6 & 5.9 & 4.3 & 5.2 & 4.8 & 4.3 & 6.1 & 5.9 \\
\addlinespace
\midrule
\addlinespace

\multirow{6}{*}{\rotatebox{90}{\textbf{\gpqa}}}
& \qwmini    & 17.4 & 19.4 & 19.9 & 20.8 & 19.4 & 19.4 & 20.5 & 19.4 & 20.1 & 21.2 & 20.8 & 21.0 & 21.0 & 16.1 & 16.7 & 16.3 & 19.6 & 20.8 & 18.1 & 18.1 & 17.2 & 14.5 & 17.2 & 18.8 & 19.4 \\
& \qwsmall   & 16.7 & 21.2 & 23.9 & 22.3 & 23.0 & 20.8 & 25.0 & 21.9 & 22.3 & 22.8 & 24.8 & 20.3 & 21.2 & 19.4 & 17.2 & 15.6 & 20.3 & 17.9 & 21.4 & 18.3 & 19.6 & 16.7 & 18.5 & 20.1 & 19.4 \\
& \qwmedium  & 15.4 & 25.4 & 28.6 & 26.8 & 27.9 & 25.0 & 25.2 & 26.8 & 29.2 & 29.7 & 25.7 & 24.6 & 26.6 & 21.2 & 21.0 & 21.0 & 21.4 & 24.1 & 24.6 & 19.9 & 20.3 & 19.2 & 26.8 & 24.6 & 23.0 \\
& \llmini    & 14.7 & 16.3 & 15.4 & 17.0 & 14.5 & 13.8 & 12.7 & 14.7 & 15.8 & 15.4 & 14.5 & 12.3 & 12.7 & 15.4 & 15.6 & 14.3 & 16.1 & 15.4 & 15.2 & 14.3 & 14.5 & 15.0 & 15.8 & 15.2 & 14.1 \\
& \llsmall   & 14.3 & 14.3 & 15.8 & 15.6 & 15.4 & 15.4 & 13.2 & 17.2 & 12.9 & 16.3 & 12.3 & 16.5 & 17.9 & 10.5 & 10.5 & 12.5 & 14.5 & 13.8 & 10.7 & 12.3 & 12.1 & 9.4 & 13.4 & 10.5 & 11.4 \\
& \mistral   & 9.6 & 5.1 & 6.2 & 5.8 & 6.0 & 7.6 & 6.2 & 6.0 & 4.9 & 6.5 & 6.5 & 7.1 & 6.0 & 8.0 & 6.7 & 4.0 & 0.9 & 1.6 & 4.5 & 5.1 & 3.3 & 4.7 & 4.0 & 2.9 & 3.8 \\
\addlinespace
\midrule
\addlinespace

\multirow{6}{*}{\rotatebox{90}{\textbf{\gsmk}}}
& \qwmini    & 81.9 & 88.0 & 88.0 & 86.0 & 88.0 & 89.0 & 87.0 & 88.5 & 88.0 & 88.0 & 88.0 & 86.5 & 88.5 & 88.0 & 89.0 & 89.0 & 88.0 & 89.0 & 88.5 & 87.5 & 86.5 & 86.0 & 87.5 & 88.5 & 88.0 \\
& \qwsmall   & 85.3 & 94.5 & 94.0 & 93.5 & 93.0 & 94.0 & 94.0 & 94.0 & 94.0 & 94.5 & 93.5 & 94.0 & 94.0 & 89.0 & 86.0 & 91.5 & 92.0 & 94.0 & 93.5 & 86.5 & 90.5 & 92.5 & 92.5 & 94.0 & 92.5 \\
& \qwmedium  & 90.4 & 96.5 & 97.0 & 97.0 & 96.5 & 97.5 & 98.0 & 96.5 & 96.5 & 97.0 & 96.5 & 95.0 & 96.5 & 94.5 & 93.0 & 94.5 & 95.0 & 95.0 & 96.0 & 95.0 & 94.5 & 95.5 & 94.5 & 96.5 & 95.0 \\
& \llmini    & 62.3 & 75.5 & 75.5 & 77.0 & 78.0 & 72.0 & 80.0 & 74.0 & 76.5 & 76.0 & 79.5 & 74.0 & 80.0 & 69.5 & 67.0 & 71.5 & 66.0 & 69.5 & 68.0 & 67.5 & 65.0 & 67.0 & 65.0 & 66.5 & 68.0 \\
& \llsmall   & 78.0 & 85.5 & 87.0 & 89.0 & 86.0 & 85.5 & 88.5 & 88.5 & 89.5 & 88.0 & 86.5 & 84.0 & 89.0 & 83.0 & 84.5 & 84.0 & 85.5 & 82.5 & 83.0 & 85.0 & 81.0 & 80.5 & 84.0 & 82.5 & 84.5 \\
& \mistral   & 5.0 & 60.0 & 32.5 & 28.0 & 28.5 & 45.5 & 38.0 & 27.0 & 28.5 & 41.0 & 42.0 & 45.5 & 30.0 & 32.0 & 12.0 & 10.5 & 4.0 & 38.5 & 21.0 & 4.0 & 39.5 & 3.5 & 39.5 & 36.0 & 33.0 \\
\addlinespace
\midrule
\addlinespace

\multirow{6}{*}{\rotatebox{90}{\textbf{\hle}}}
& \qwmini    & 2.0 & 1.6 & 1.6 & 1.9 & 1.7 & 1.7 & 1.7 & 1.9 & 1.7 & 1.6 & 1.6 & 1.9 & 1.7 & 2.4 & 2.1 & 1.9 & 2.5 & 2.4 & 2.2 & 1.9 & 2.4 & 2.3 & 2.3 & 2.2 & 2.5 \\
& \qwsmall   & 2.0 & 3.0 & 3.1 & 2.9 & 2.9 & 2.8 & 2.9 & 2.6 & 2.7 & 2.6 & 2.9 & 3.0 & 2.4 & 2.4 & 1.9 & 2.1 & 2.5 & 2.2 & 2.6 & 1.9 & 2.6 & 2.2 & 2.5 & 2.1 & 2.3 \\
& \qwmedium  & 1.9 & 1.7 & 1.9 & 1.7 & 2.2 & 1.5 & 1.7 & 1.8 & 1.8 & 1.7 & 1.6 & 1.7 & 1.6 & 2.1 & 2.4 & 2.7 & 2.2 & 1.7 & 2.5 & 2.4 & 1.9 & 2.4 & 2.7 & 1.9 & 2.8 \\
& \llmini    & 0.3 & 2.3 & 2.7 & 2.2 & 2.0 & 2.4 & 2.3 & 2.0 & 2.4 & 1.9 & 2.2 & 2.5 & 2.3 & 1.5 & 1.3 & 1.4 & 1.1 & 1.1 & 1.2 & 1.1 & 0.8 & 1.1 & 1.1 & 1.8 & 1.4 \\
& \llsmall   & 2.2 & 2.4 & 2.3 & 2.1 & 2.0 & 1.9 & 2.2 & 2.7 & 2.7 & 2.7 & 2.3 & 2.4 & 2.0 & 2.6 & 2.7 & 2.5 & 2.2 & 2.5 & 2.1 & 2.2 & 2.4 & 2.1 & 2.1 & 2.3 & 2.8 \\
& \mistral   & 0.3 & 0.7 & 1.1 & 0.8 & 0.8 & 1.7 & 0.7 & 0.9 & 0.6 & 0.9 & 0.8 & 1.3 & 0.9 & 1.2 & 1.1 & 0.9 & 0.3 & 0.8 & 0.9 & 0.6 & 0.9 & 0.6 & 0.6 & 0.5 & 1.0 \\
\addlinespace
\midrule
\addlinespace

\multirow{6}{*}{\rotatebox{90}{\textbf{\mathfour}}}
& \qwmini    & 57.4 & 59.4 & 60.6 & 59.6 & 58.6 & 59.1 & 58.6 & 59.6 & 59.9 & 59.1 & 59.4 & 58.6 & 59.1 & 58.6 & 57.4 & 60.3 & 60.1 & 60.1 & 58.6 & 58.4 & 58.4 & 58.1 & 58.1 & 59.4 & 58.6 \\
& \qwsmall   & 57.9 & 61.3 & 61.3 & 61.3 & 61.3 & 61.1 & 61.6 & 61.6 & 62.3 & 61.3 & 60.6 & 62.3 & 62.6 & 58.9 & 57.4 & 61.6 & 61.8 & 62.3 & 60.1 & 61.3 & 58.6 & 61.1 & 60.1 & 64.1 & 62.1 \\
& \qwmedium  & 64.1 & 64.3 & 65.1 & 62.8 & 61.3 & 60.1 & 61.1 & 65.1 & 63.6 & 65.3 & 62.3 & 60.6 & 62.8 & 67.3 & 67.8 & 67.1 & 66.3 & 67.8 & 67.6 & 67.1 & 66.8 & 67.1 & 66.8 & 67.8 & 68.3 \\
& \llmini    & 45.9 & 57.6 & 56.9 & 56.4 & 55.9 & 55.6 & 56.9 & 57.9 & 57.6 & 56.4 & 55.6 & 55.9 & 55.1 & 49.9 & 50.1 & 49.4 & 51.4 & 48.6 & 50.6 & 50.4 & 48.1 & 52.9 & 51.1 & 47.1 & 50.6 \\
& \llsmall   & 50.6 & 53.9 & 53.4 & 53.6 & 52.9 & 53.1 & 51.6 & 53.6 & 52.4 & 51.9 & 53.6 & 51.9 & 54.4 & 52.6 & 52.6 & 50.6 & 50.4 & 48.6 & 51.4 & 50.9 & 51.4 & 50.6 & 47.9 & 48.6 & 51.9 \\
& \mistral   & 0.0 & 9.0 & 8.0 & 7.7 & 7.0 & 11.5 & 6.2 & 8.5 & 7.0 & 8.0 & 7.7 & 10.0 & 7.0 & 11.7 & 9.2 & 5.2 & 7.0 & 7.0 & 1.7 & 28.4 & 8.7 & 0.5 & 14.7 & 8.7 & 3.7 \\
\addlinespace
\midrule
\addlinespace

\multirow{6}{*}{\rotatebox{90}{\textbf{\mathfive}}}
& \qwmini    & 52.8 & 50.8 & 51.2 & 49.4 & 50.8 & 52.4 & 51.0 & 51.0 & 51.0 & 51.2 & 51.4 & 51.0 & 51.4 & 49.8 & 50.6 & 51.8 & 51.8 & 51.6 & 51.2 & 52.4 & 51.6 & 52.0 & 51.8 & 52.2 & 51.0 \\
& \qwsmall   & 46.8 & 60.6 & 60.2 & 61.2 & 60.0 & 58.8 & 60.2 & 59.4 & 60.0 & 59.6 & 61.0 & 59.8 & 61.2 & 54.0 & 57.6 & 59.0 & 60.2 & 60.4 & 60.0 & 55.0 & 56.8 & 61.0 & 61.4 & 60.6 & 60.2 \\
& \qwmedium  & 61.6 & 63.4 & 63.2 & 63.0 & 63.8 & 64.6 & 62.0 & 63.0 & 64.0 & 63.6 & 64.6 & 64.8 & 63.0 & 63.2 & 62.0 & 62.8 & 62.2 & 64.4 & 63.4 & 60.6 & 63.0 & 62.8 & 64.4 & 63.2 & 62.6 \\
& \llmini    & 26.6 & 33.2 & 32.6 & 32.0 & 32.6 & 34.4 & 33.4 & 31.4 & 32.0 & 36.6 & 34.2 & 33.2 & 32.0 & 32.4 & 30.2 & 32.6 & 34.4 & 34.8 & 31.8 & 31.2 & 32.6 & 31.0 & 34.8 & 33.2 & 32.6 \\
& \llsmall   & 33.8 & 39.0 & 38.0 & 37.6 & 38.4 & 37.0 & 37.2 & 37.8 & 37.2 & 37.2 & 37.2 & 36.8 & 36.6 & 38.2 & 35.6 & 35.8 & 36.6 & 33.0 & 36.2 & 34.6 & 35.6 & 36.4 & 36.6 & 36.2 & 35.8 \\
& \mistral   & 2.2 & 8.8 & 9.8 & 9.8 & 8.4 & 10.8 & 13.0 & 7.8 & 10.0 & 10.6 & 10.2 & 9.2 & 9.0 & 9.8 & 8.4 & 6.8 & 6.0 & 10.0 & 8.4 & 4.0 & 11.2 & 4.4 & 12.6 & 10.8 & 10.8 \\
\addlinespace
\bottomrule
\end{tabular}
}
\end{table*}


\section{Correlation Calculation Details}
\label{app:correlation}

To rigorously quantify the relationship between training code data complexity and model reasoning performance, we employ Spearman's rank correlation coefficient ($\rho$)~\cite{dodgeSpearmanRankCorrelation2008}.
This non-parametric measure is chosen because our complexity levels (\textsc{min}, \lowspl, \midspl, \highspl, \maxspl). represent an ordinal scale rather than a continuous ratio scale, making Pearson's correlation less appropriate.
Specifically, for each base model and dataset combination, we calculate $\rho$ between the integer-mapped difficulty levels ($0-4$) and the corresponding evaluation accuracy.
The calculation is performed using the \texttt{scipy.stats.spearmanr} function from the \texttt{SciPy}\footnote{SciPy: \href{https://scipy.org/}{https://scipy.org/}} library, which also provides two-sided $p$-values to assess statistical significance.
A standard significance level of $\alpha = 0.05$ is used to determine if the observed correlations are statistically significant.
We exclude control groups from this specific analysis to focus solely on the trend within the complexity-stratified fine-tuning runs.


\end{document}